\documentclass{article}
\usepackage{microtype}
\usepackage{graphicx} 
\usepackage{subfigure}
\usepackage{booktabs} 

\usepackage{hyperref}

\usepackage[utf8]{inputenc} 
\usepackage[T1]{fontenc}    
\usepackage{booktabs}       
\usepackage{amsfonts}       
\usepackage{nicefrac}       
\usepackage{microtype}      
\usepackage{xcolor}         
\usepackage{bm}
\usepackage{arydshln}

\usepackage[accepted]{icml2024_arxiv}

\usepackage{amsmath}
\usepackage{amssymb}
\usepackage{mathtools}
\usepackage{amsthm}

\usepackage[capitalize,noabbrev]{cleveref}

\theoremstyle{plain}
\newtheorem{thm}{Theorem}[section]
\newtheorem{lem}[thm]{Lemma}
\newtheorem{prop}[thm]{Proposition}
\newtheorem{hyp}[thm]{Hypothesis}
\newtheorem{col}[thm]{Corollary}

\theoremstyle{definition}
\newtheorem{defi}[thm]{Definition}

\usepackage[textsize=tiny]{todonotes}

\usepackage{wrapfig}

\usepackage{amsmath,amsfonts,bm}

\def\eqref#1{equation~\ref{#1}}

\def\1{\bm{1}}

\DeclareMathAlphabet{\mathsfit}{\encodingdefault}{\sfdefault}{m}{sl}
\SetMathAlphabet{\mathsfit}{bold}{\encodingdefault}{\sfdefault}{bx}{n}

\newcommand{\R}{\mathbb{R}}

\DeclareMathOperator{\Tr}{Tr}

\DeclareMathOperator{\LN}{LN}
\usepackage{comment}
\usepackage{xcolor}

\newcommand{\N}{\mathbb{N}}

\newcommand{\arxiv}[1]{{#1}}

\icmltitlerunning{Understanding MLP-Mixer as a Wide and Sparse MLP}

\begin{document}

\twocolumn[
\icmltitle{Understanding MLP-Mixer as a Wide and Sparse MLP}



\icmlsetsymbol{equal}{*}

\begin{icmlauthorlist}
\icmlauthor{Tomohiro Hayase}{xxx}
\icmlauthor{Ryo Karakida}{yyy}
\end{icmlauthorlist}

\icmlaffiliation{yyy}{Artificial Intelligence Research Center, AIST}
\icmlaffiliation{xxx}{Metaverse Lab, Cluster Inc.}

\icmlcorrespondingauthor{Tomohiro Hayase}{t.hayase@cluster.mu}
\icmlcorrespondingauthor{Ryo Karakida}{karakida.ryo@aist.go.jp}


\vskip 0.3in
]



\printAffiliationsAndNotice{Preprint of a paper to appear in ICML 2024.}  

\begin{abstract}
Multi-layer perceptron (MLP) is a fundamental component of deep learning, and recent MLP-based architectures, especially the MLP-Mixer, have achieved significant empirical success. Nevertheless, our understanding of why and how the MLP-Mixer outperforms conventional MLPs remains largely unexplored. In this work, we reveal that sparseness is a key mechanism underlying the MLP-Mixers. First, the Mixers have an effective expression as a wider MLP with Kronecker-product weights, clarifying that the Mixers efficiently embody several sparseness properties explored in deep learning. In the case of linear layers, the effective expression elucidates an implicit sparse regularization caused by the model architecture and a hidden relation to Monarch matrices, which is also known as another form of sparse parameterization. Next, for general cases, we empirically demonstrate quantitative similarities between the Mixer and the unstructured sparse-weight MLPs. Following a guiding principle proposed by Golubeva, Neyshabur and Gur-Ari (2021), which fixes the number of connections and increases the width and sparsity, the Mixers can demonstrate improved performance.

\end{abstract}

\section{Introduction}
Multi-layer perceptron (MLP) and its variants are fundamental components of deep learning employed in various problems and for understanding the basic properties of neural networks.
Despite their simplicity and long history \citep{rosenblatt1958perceptron,schmidhuber2015deep}, 
it has become apparent only recently that there is still significant room for improvement in the predictive performance of MLP-based architectures.

The sparseness is known as a key direction for enhancing the performance of dense MLP layers~\citep{neyshabur2014search,d2019finding,neyshabur2020towards,golubeva2021,pellegrini2022neural}. For instance, 
\citet{golubeva2021} reported that the prediction performance improves by increasing both the width and the sparsity of connectivity when the number of trainable parameters is fixed. 
The MLP-Mixer is another noteworthy direction of recent developments in MLP-based architectures~\citep{Tolstikhin21mlp-mixer,touvron2022resmlp}.
It does not rely on convolutions or self-attention and is entirely composed of MLP layers; instead, it utilizes MLPs applied across spatial locations or feature channels. This can be regarded as an MLP that applies fully connected (FC) layers on both sides of the feature matrix. Despite its simplicity, the MLP-Mixer achieved a performance on image classification benchmarks comparable to that of more structured deep neural networks.

However, there are few studies on experimental and theoretical attempts to understand MLP-Mixer's internal mechanisms \citep{yu2022metaformer,sahiner2022unraveling}. This contrasts with the extensive elucidation of explicit or implicit biases in other modern architectural components \cite{neyshabur2014search,bjorck2018understanding,cordonnier2019relationship}.
To further advance the MLP-based architecture, it will be crucial to unveil the underlying mechanism of the MLP-Mixer and to address questions such as: What different inductive biases does the MLP-Mixer have compared to a naive MLP? Which architectural factors significantly contribute to their superior performance? 

In this study, we reveal that the sparseness, which is seemingly a distinct research concept, is
the key mechanism underlying the MLP-Mixer. 
One can see that the MLP-Mixer is a wider MLP with sparsity inherently embedded as an inductive bias. This equivalence also provides a quantitative understanding that appropriate token and channel sizes maximizing the sparsity can improve the prediction performance.
The detailed contributions are summarized as follows: 

\begin{itemize}

\item  We first identify {\it an effective expression of MLP-Mixer as an MLP} by vectorizing the mixing layers (in Section \ref{sec:similarity}). It is composed of the permutation matrix and the Kronecker product and provides an interpretation of mixing layers as an extremely wide MLP with sparse (structured) weights.   

\item 
We consider a linear activation case of a simple Mixer and its effective expression. 
First, this reveals that the Kronecker-product weight possesses a bias towards an implicit L1 regularization (in Section \ref{sec:regularization}).
In deep learning, there are two aspects of sparsity: one is a large number of zero parameters (e.g., \citet{neyshabur2020towards}) and the other is the limited number of independent parameters (e.g., \citet{dao2022monarch}).  Our evaluation of implicit regularization effectively links these two facets of sparsity.
Second, regarding the limited number of independent parameters, 
the MLP-Mixer can be regarded as an approximation of an MLP with the Monarch matrix (in Section \ref{sec:monarch}).
Thus, the sparseness discussed in the different literature is inherently and originally incorporated into the Mixers.

\item For realistic cases with non-linear activation functions, we quantitatively evaluate the high similarity between the traditional sparse-weight MLP (SW-MLP) and the MLP-Mixer. 
First, we confirm the similarity of hidden features by the centered kernel alignment.
 Second, we reveal similar performance trends when increasing sparseness (equivalent to widening) with a fixed number of connections. This means that empirical observations of \citet{golubeva2021} on appropriate sizes of sparsity hold even in the MLP-Mixer in the sense of both improving prediction performance (in Section \ref{sec:acc_vs_sw}) and ensuring trainability (in Section \ref{sec:spectrum}). We also empirically verify that the Random-Permuted (RP) Mixer introduced in Section \ref{sec:rp-mixer}, which is a less structured variant of the normal Mixer, exhibits similar performance trends (Sections \ref{sec:experiments-2} \& \ref{sec:experiments-1}). This further solidifies our understanding that sparsity is a fundamental element underlying the Mixers.

\end{itemize}

\section{Preliminaries}

\subsection{Related Work}

\noindent{\bf MLP-based architectures.}
  The salient property of an MLP-Mixer is that it is composed entirely of FC layers. This property is unique to the MLP-Mixer (and its concurrent work ResMLP \citep{touvron2022resmlp}) and different from attention-based architectures \citep{Dosovitskiy21vit}. 
While some previous work focused on providing a relative evaluation of performance compared with the attention module  \citep{yu2022metaformer,sahiner2022unraveling}, our purpose is to elucidate the hidden bias of MLP-Mixers as a wide and sparse MLP.
 \citet{golubeva2021} investigated that the generalization performance can be improved by increasing the width in FC layers. Because they fixed the number of weights, an increase in the width caused a higher sparsity. They revealed that even for fixed sparse connectivity throughout training, a large width can 
improve the performance better than the dense layer. 

\paragraph{Structured weight matrices: (i) Sparse matrix.}
Parameter sparsity is widely used to improve the performance and efficiency of neural networks. 
It is known that naive MLPs require even more data than structured deep models to improve performance due to their weak inductive bias \cite{bachmann2023scaling}. The sparseness is considered useful as a minimal necessary bias.
One approach to make weights sparse is to determine nonzero weights dynamically, such as dense-to-sparse training \citep{neyshabur2020towards}, pruning \citep{frankle2018the}, and sparse-to-sparse training \citep{dettmers2019sparse,evci2020rigging}.
The other is to constrain the trainable weights from the beginning of training statically \citep{dao2022monarch,golubeva2021,liu2022unreasonable,gadhikar2023random}. The current study follows the latter approach; specifically, we reveal that the mixing layers of the MLP-Mixer are implicitly related to such fixed sparse connectivity. 
{\bf (ii) Kronecker product.} 
Constraining weight matrices to the Kronecker product and its summation has been investigated in the model-compression literature. Some works succeeded in reducing the number of trainable parameters without deteriorating the prediction performance \citep{zhou2015exploiting,zhangbeyond}
while others applied them for the compression of trained parameters \citep{hameed2022convolutional}.
In contrast, we find a Kronecker product expression hidden in the MLP-Mixer, which can be regarded as an approximation of the Monarch matrix proposed in \citet{dao2022monarch}.

Notably, our study is completely different from merely applying sparse regularizers, Kronecker-product weights, or Monarch matrices  \cite{dao2022monarch} to the dense weight matrices in the mixing layers like \citet{fu2023monarch}. Our finding is that MLP-Mixers and their generalization (i.e., the PK family) inherently possess these properties.

\subsection{Notations}

\paragraph{MLP-Mixer.}
An MLP-Mixer is defined as follows \citep{Tolstikhin21mlp-mixer}. Initially, it divides an input image into patches. Next, a per-patch FC layer is performed. After that, the blocks described as follows are repeatedly applied to them: for the feature matrix from the previous hidden layer $X\in\mathbb{R}^{S\times C}$,   
\begin{align}
\text{Token-MLP}(X) &=  W_2 \phi(W_1 X), \\ \text{Channel-MLP}(X)
&=  \phi( X W_3) W_4,
\label{eq6:0428}
\end{align}
where $\phi$ denotes the entry-wise activation function,   $W_1 \in \mathbb{R}^{\gamma S\times S}$, $W_2 \in \mathbb{R}^{S \times \gamma S}$, $W_3 \in \mathbb{R}^{C\times \gamma C}$ and $W_4 \in \mathbb{R}^{\gamma C\times C}$. In this paper, we set the expansion factor of the hidden layers of  token and channel-mixing MLPs to the same value $\gamma$ for simplicity. 
The block of the MLP-Mixer is given by the map $X \mapsto Y$, where
\begin{align}
U &= X + \text{Token-MLP}(\text{LN}(X)), \\ 
Y &= U + \text{Channel-MLP}(\text{LN}(U)). \label{eq:mixer}
\end{align}
In the end, the global average pooling and the linear classifier are applied to the last hidden layer.

\paragraph{S-Mixer.}

To facilitate theoretical insights, we introduce the S-Mixer, an idealized version of the MLP Mixer:
\begin{align}\label{align:feature}
H = \phi (\phi(WX)V),
\end{align}
where $V$ and $W$ represent weight matrices. In this formulation, the shallow neural networks typically found within the mixing block of the MLP-Mixer are simplified to single layers. For the sake of simplicity in our theoretical analysis of the S-Mixer, we omit layer normalization and skip connections. It is important to note, however, that in our numerical experiments on training deep models in \cref{sec:beyond}, these components are incorporated, demonstrating that their inclusion does not detract from the fundamental outcomes of the study. we summarized the detailed architectures used for experiments in \cref{sec:arch}.

\section{Properties as a sparse MLP}\label{sec:similarity}

\subsection{Vectorization}
\begin{figure*}[t]
    \centering
    \includegraphics[width=0.99\linewidth]{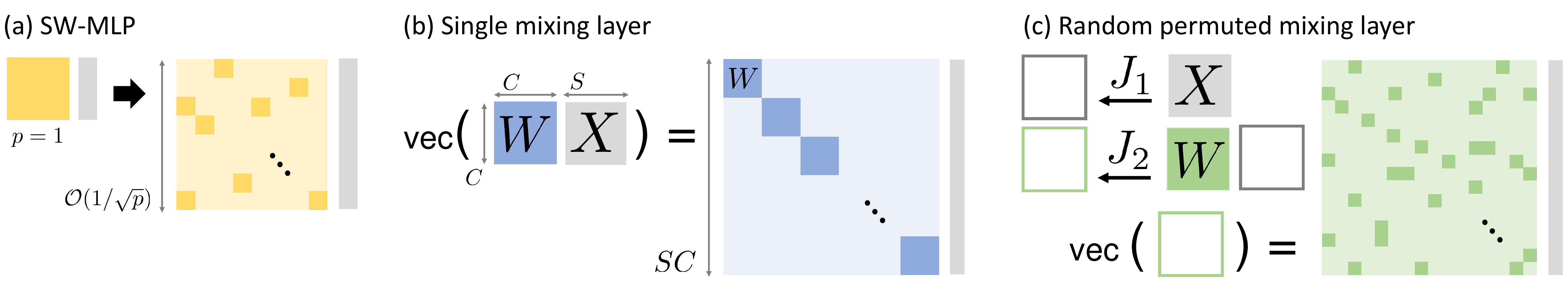}
    \caption{Schematic diagram of sparsity treated in this work.  (a) A masked weight matrix $M \odot A$ in a sparse-weight MLP (SW-MLP).  Its width is  $O(1/\sqrt{p})$, where $p$ is the ratio of non-zero entries in the mask $M$. (b) A mixing layer in an MLP-Mixer with the vectorization.  The weight behaves as a block diagonal matrix. (c) A weight of a random permuted mixer (RP-Mixer), which is introduced in \cref{sec:beyond}. The block diagonal structure is destroyed by random permutation matrices  $J_1,J_2$ to achieve similarity to the SW-MLP.}
    \label{fig:schematic}
\end{figure*}

To address the similarity between MLP-Mixer and MLP, we consider vectorization of feature tensors and effective width.
We represent the vectorization operation of the matrix $X \in \mathbb{R}^{S \times C}$ by $\text{vec}(X)$; more precisely, $(\text{vec}(X))_{{ (j-1)d + i}} =  X_{ij} , (i = 1, \dots, S,  j= 1, \dots, C)$. We also define an inverse operation $\text{mat}(\cdot)$ to recover the matrix representation by $\text{mat}(\text{vec}(X)) = X$.
There exists a well-known equation for the vectorization operation  and the Kronecker product denoted by $\otimes$; 
\begin{equation}
  \text{vec}(W X V) =  (V^\top \otimes W) \text{vec}(X), \label{eq:wxv}
\end{equation}
for $W \in \mathbb{R}^{S \times S}$ and $V \in \mathbb{R}^{C \times C}$. 
The vectorization of the feature matrix $W X V$ is equivalent to a fully connected layer of width $m:=SC$ with a weight matrix  $V^\top \otimes W$. We refer to this $m$ as the {\it effective width} of mixing layers.

In MLP-Mixer, when we treat each $S \times C$ feature matrix $X$ as an  $SC$-dimensional vector $\mathrm{vec}(X)$, the right multiplication by an $C \times C$ weight $V$ and  the left weight multiplication by a $S \times S$ weight  $W$ are represented as follows:
\begin{align}
\mathrm{vec}(XV) &= (V^\top \otimes I_S ) \mathrm{vec}(X),\\ \mathrm{vec}(WX) &= (I_C \otimes W)\mathrm{vec}(X)
\end{align}
where  $I_n$ denotes an $n \times n$ identity matrix.
This expression clarifies that the mixing layers work as an MLP with special weight matrices with the Kronecker product. As usual, the size of $S$ and $C$ is approximately $10^2 \sim 10^3$, and this implies that the Mixer is equivalent to an extremely wide MLP with $m = 10^4 \sim 10^6$. Moreover, the ratio of non-zero entries in the weight matrix $I_C \otimes W$ is $1/C$ and that of $V^\top \otimes I_S$ is $1/S$. 
Therefore, the weight of the effective MLP is highly sparse in the sense of non-zero entries.

Here, to consider only the left-multiplication of weights, we introduce commutation matrices.
\begin{defi}
   The commutation matrix $J_c$  is an $m \times m$ matrix defined as
\begin{equation}
J_c \text{vec}(X) =  \text{vec}(X^\top),
\end{equation}
where $X$ is an $S \times C$ matrix.  
\end{defi}
Note that for any $x \in \R^m$,  
     $J_c \phi (x)  = \phi( J_cx).$
In addition, we have  $J_c^\top (I_S \otimes V^\top) J_c = V^\top \otimes I_S$ for any $C \times C$ matrix $V$. 
Using the commutation matrix, we find the following:
\begin{prop}[Effective expression of MLP-Mixer as MLP]
The feature matrix of the S-Mixer (\ref{align:feature}) is a shallow MLP with width $m=SC$ as follows: 
\begin{align}
\text{vec}(H) &=  \phi \left(J_c^\top \left(  I_S \otimes V^\top\right)   \phi \left(J_c \left(I_C \otimes W\right) \text{vec}\left(X\right)\right)\right). \label{eq11:0502} 
\end{align}
\end{prop}
The derivation is straightforward as described in \cref{sec:proof_expression}. 
This expression clarifies that the mixing layers work as an MLP with special weight matrices with the commutation matrix and Kronecker product.

It is easy to generalize the above expression for the S-Mixer to the MLP-Mixer, where each mixing operation is composed of shallow neural networks (\ref{eq6:0428}) (see \cref{sec:proof_expression}).
This equivalence with a wide MLP with sparse weights is simple and easy to follow but has been missing in the literature.

\subsection{Implicit regularization of linear mixing layers}\label{sec:regularization}
Considering the linear activation case to gain theoretical insight is a common approach used in deep learning  \cite{saxe2014exact,arora2019implicit}. yet there has been no work on the Mixers. We find that the theoretical analysis is challenging even in the simplest case of the linear S-Mixer (\ref{eq:wxv}). This model is equivalent to a so-called bi-linear regression model, for which an analytical solution that minimizes the MSE loss is unknown \cite{hoff2015multilinear}. 
This makes analyzing the linear S-Mixer more difficult compared to conventional linear networks. Despite this difficulty, we discover the following inequality that characterizes the implicit regularization of the model:
\begin{prop}\label{prop:regularization}
  \begin{align}
 &\min_{V,W }\mathcal{L}(V\otimes W) + \frac{\lambda}{2} (\|V\|_F^2+\|W\|_F^2) \nonumber \\
 &\geq 
\min_{B \in \mathbb{R}^{SC\times SC} }\mathcal{L}(B) + \tilde{\lambda} \|B\|_1 \ \text{with} \ \tilde{\lambda}=\lambda/CS
\end{align}
where $\|\cdot \|_F$ is the Frobenius norm, $\|\cdot \|_1$ is the L1 norm.
\end{prop}
The derivation is shown in \cref{sec:proof_regularization}. It extends the fact that the Hadamard product parameterization has an implicit bias towards L1 regularization to the case of the Kronecker product parameterization \cite{hoff2017lasso,yasuda2022sequential}. 
Note that the number of independent parameters differs between the left and right sides of the inequality. Therefore, $\tilde{\lambda}$ might appear small, but it merely normalizes the change in parameter size~(see \cref{sec:proof_regularization} for more details). 
The implicit bias towards sparsity is considered a desirable property in modern neural network architectures \cite{neyshabur2014search,woodworth2020kernel}.

\subsection{Monarch matrix hidden behind Mixers}\label{sec:monarch}

\begin{figure*}[th]
    \centering
    \includegraphics[width=0.99\linewidth]{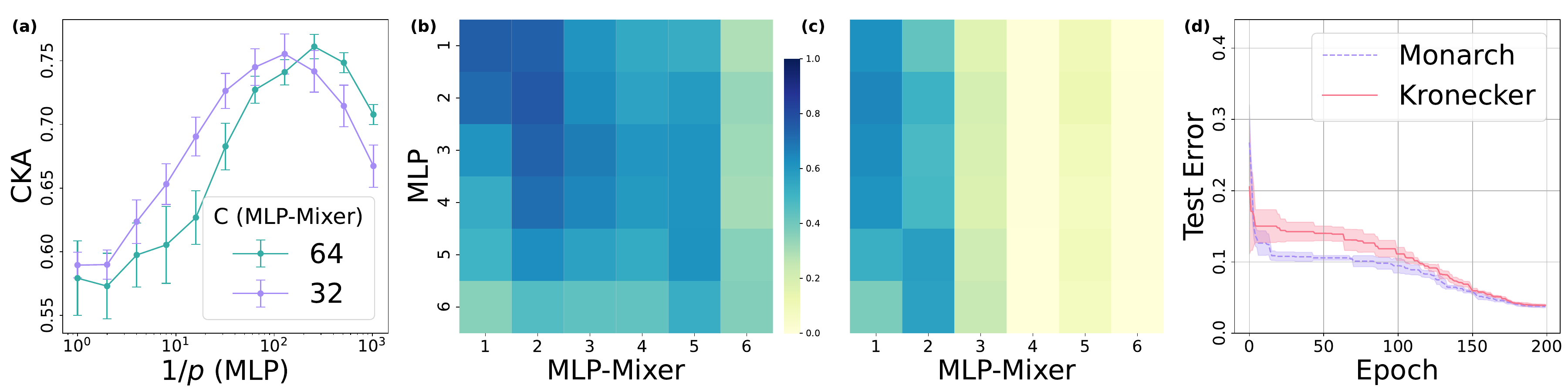}
    \caption{(a) Average of diagonal entries of CKA between trained MLP-Mixer ($S=C=64,32$) and MLP with different sparsity,  where $p$ is the ratio of non-zero entries in $M$. (b) CKA between MLP-Mixer ($S=C=64$) and MLP with the corresponding $p=1/64$, and (c) CKA between the Mixer and a dense MLP.  (d) Test error on MNIST of shallow MLPs with Monarch matrix weights and Kronecker weights. The result is the average of five trials with different random seeds.}
    \label{fig:cka_fr_acc}
\end{figure*}

\citet{dao2022monarch} proposed a {\it Monarch matrix $M \in \mathbb{R}^{n \times n}$} defined by 
\begin{equation}
M=J_c^\top L J_c R,    \label{monarch}
\end{equation}
 where $L$ and $R$ are the trainable block diagonal matrices, each with $\sqrt{n}$ blocks of size $\sqrt{n} \times \sqrt{n}$. The previous work claimed that the Monarch matrix is sparse in that the number of trainable parameters is much smaller than in a dense $n \times n$ matrix.    
Despite this sparsity, by replacing the dense matrix with a Monarch matrix, it was found that various architectures can achieve almost comparable performance while succeeding in shortening the training time. Furthermore, the product of a few Monarch matrices can represent many commonly used structured matrices such as convolutions and Fourier transformations.

Surprisingly, the MLP Mixer and the Monarch matrix, two completely different concepts, have hidden connections. 
By comparing (\ref{eq11:0502}) and (\ref{monarch}), we find that 
\begin{col}\label{col:monarch}
 Consider a S-Mixer without an intermediate activation function, that is,  its feature matrix is given by $H=\phi(WXV)$. Then, $\text{vec}(H)$  is equivalent to an MLP whose weight matrix is given by a Monarch matrix with weight-sharing diagonal matrices, that is, $\phi(Mx)$ with $n=SC$, $L = I_S \otimes V^\top$ and $R=I_C \otimes W$. 
\end{col}

We validate the similarities between the Monarch matrix and the Mixer through experiments. Here, we consider a shallow MLP (\(x \mapsto B_c B_2 \mathrm{ReLU}(B_1x)\)) where \(B_c\) is the classification layer. We compare the performance of models where weights \(B_i\) (\(i=1,2\)) are replaced with Monarch matrices \(M_i\), and models where they are replaced with Kronecker products \(V_i \otimes W_i\). In \cref{fig:cka_fr_acc}~(left), the two models showed comparable performance. This suggests that although the Mixer incorporates a weight-sharing structure compared to Monarch, its performance is not compromised.

\subsection{Comparing  hidden features}\label{sec:cka}

Let us back to the situation in practice where the model has intermediate non-linear activation. 
To investigate the similarity of non-linear networks, here we consider an  \emph{unstructured sparse-weight MLP} (SW-MLP in short), which is a basic implementation in the literature of naive MLPs with sparse weights~\citep{golubeva2021}.
Its weights are given by a random sparse mask, that is,  $M \odot A$ where $A$ is the original weight matrix and $M$ is a static mask matrix whose entries are drawn from the Bernoulli distribution with a probability $p>0$  of being one at the initialization phase. 
In this section, to compare the MLP-Mixer with SW-MLP sharing conditions, we consider the average $p=(S^{-1}+C^{-1})/2$ of the sparsity of MLP-Mixer for the setting of SW-MLP.
\cref{fig:schematic} overviews the models that we compare in the below. The random permuted one is an alternative to SW-MLP
and shows the result in \cref{sec:beyond}.

As a similarity measure, 
we use the centered kernel alignment (CKA) \citep{nguyen2021do} between hidden features of MLPs with sparse weights and those of MLP-Mixers. In practice, we computed the mini-batch CKA \citep[Section~3.1(2)]{nguyen2021do} among features of trained networks. In \cref{fig:cka_fr_acc}(a), we observed the averaged CKA achieved the maximum as an appropriate sparsity of the MLPs. By comparing \cref{fig:cka_fr_acc}(b) and (c),  we found that CKA matrix with sparse MLP was clearly higher than dense MLP.  In particular, the sparse Mixer was similar to sparser MLP in hidden features. Detailed settings of all experiments are summarized in \cref{sec:experimental_setting}.

\section{Properties as a wide MLP}\label{sec:wide_mlp}
In this section, we continue the discussion on whether the MLP-Mixer (or S-Mixer) exhibits tendencies similar to those of sparse-weight MLPs. In particular, we search for desirable sparsity and discuss whether there is a desirable setting of hyper-parameters on the Mixers.

\subsection{Maximizing sparseness}
The following hypothesis has a fundamental role: 
\begin{hyp}[\citet{golubeva2021}]\label{hyp:gol}
Increasing the width up to a certain point, while keeping the number of weight parameters fixed, results in improved test accuracy.
 \label{hyp}
\end{hyp}

Intuitively, \citet{golubeva2021} challenged the question of whether the performance improvement of large-scale deep neural networks was due to an increase in the number of parameters or an increase in width.
\begin{wrapfigure}{rb}{0.23\textwidth}
\includegraphics[width=0.23\textwidth] {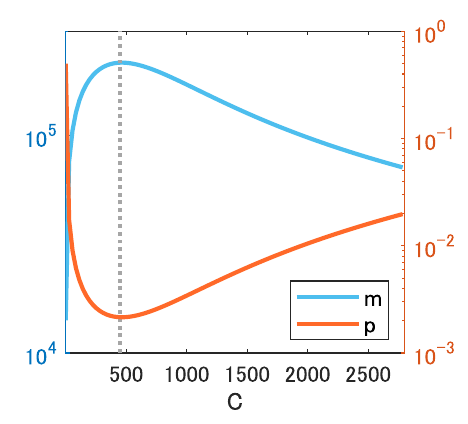}
\vspace{-12pt}
\caption{Theoretical line of $m$ and $p$ ($\Omega=10^8, \gamma=1$).}
\vspace{-0.12in}
\vspace{1pt}
\label{fig:e}
\end{wrapfigure}
They empirically succeeded in verifying Hypothesis~\ref{hyp:gol}; that is, the improvement was due to the increase in width in normal MLPs and ResNets (note that the width of ResNet indicates the channel size).

Let us denote by $\Omega$ the average number of connections per layer.
We have 
\begin{align}\label{align:omega-p}
    \Omega = p\gamma m^2,
\end{align}
where $p$ is the ratio of non-zero entries in the static mask.
Here, for the MLP-Mixer,
\begin{align}
  \Omega  = { \gamma( CS^2 + C^2S)/2}. \label{eq:omega}
\end{align}

The average number $\Omega$ of S-Mixer is reduced to $\gamma=1$ in (\ref{eq:omega}), which maintains the readability of the equations.

By (\ref{eq:omega}), we have $S ={  (\sqrt{ C^2 +  8 \Omega/(\gamma C)} - C)/ 2}$.
For a fixed $\Omega$ and $\gamma$, the effective width is controlled by $m=SC$. 
\cref{fig:e} shows $m$ as a function of $C$. The width $m$ has a single-peak form and is maximized at ($C^*,S^*$) as follows:
\begin{align}
     C^* = S^* =   (\Omega/\gamma)^{1/3}, \quad
  \max_{S,C}  m = (\Omega/\gamma)^{2/3}. \label{max_width}
\end{align}
The ratio of non-zero entries $p=\Omega/\gamma m^2$ is minimized at this point, that is, the sparsity is maximized.

\subsection{Comparing accuracy with effective width}\label{sec:acc_vs_sw}

To validate the similarity,  we compare the test error of both networks with different sparsity. Under the fixed number $\Omega$ of connectivity per layer, the sparsity is equivalent to the wideness.
\cref{fig:acc-spectrum}~(left) shows the test errors of MLP-Mixers and corresponding sparse weight MLPs under fixed $\Omega=2^{19}$ and $\gamma=2$, for several widths $\gamma m$. We observed both networks' test error improved as the width increased.   
In this sense, MLP and MLP-Mixer have a similar tendency for performance with increasing width. However,  we observed for too-wide cases around $\gamma m=8000$ in \cref{fig:acc-spectrum}~(left), the test error of SW-MLP is higher than MLP-Mixer and there is little change in response to increasing width. We discuss this tendency in the next section.

\subsection{Spectral analysis with increasing width}\label{sec:spectrum}

\citet{golubeva2021}  reported that if the sparsity became too high, the generalization performance of SW-MLP slightly decreased. They discussed that this decrease was caused by the deterioration of trainability, that is, it became difficult for the gradient descent to decrease the loss function.
Some previous work reported that the large singular values of weight matrices at random initialization cause the deterioration of trainability in deep networks \cite{bjorck2018understanding}. 
Therefore, we discuss the difference of singular values of weights between MLP-Mixer and SW-MLP.

\begin{figure}[t]
    \centering
    \includegraphics[width=0.9\linewidth]{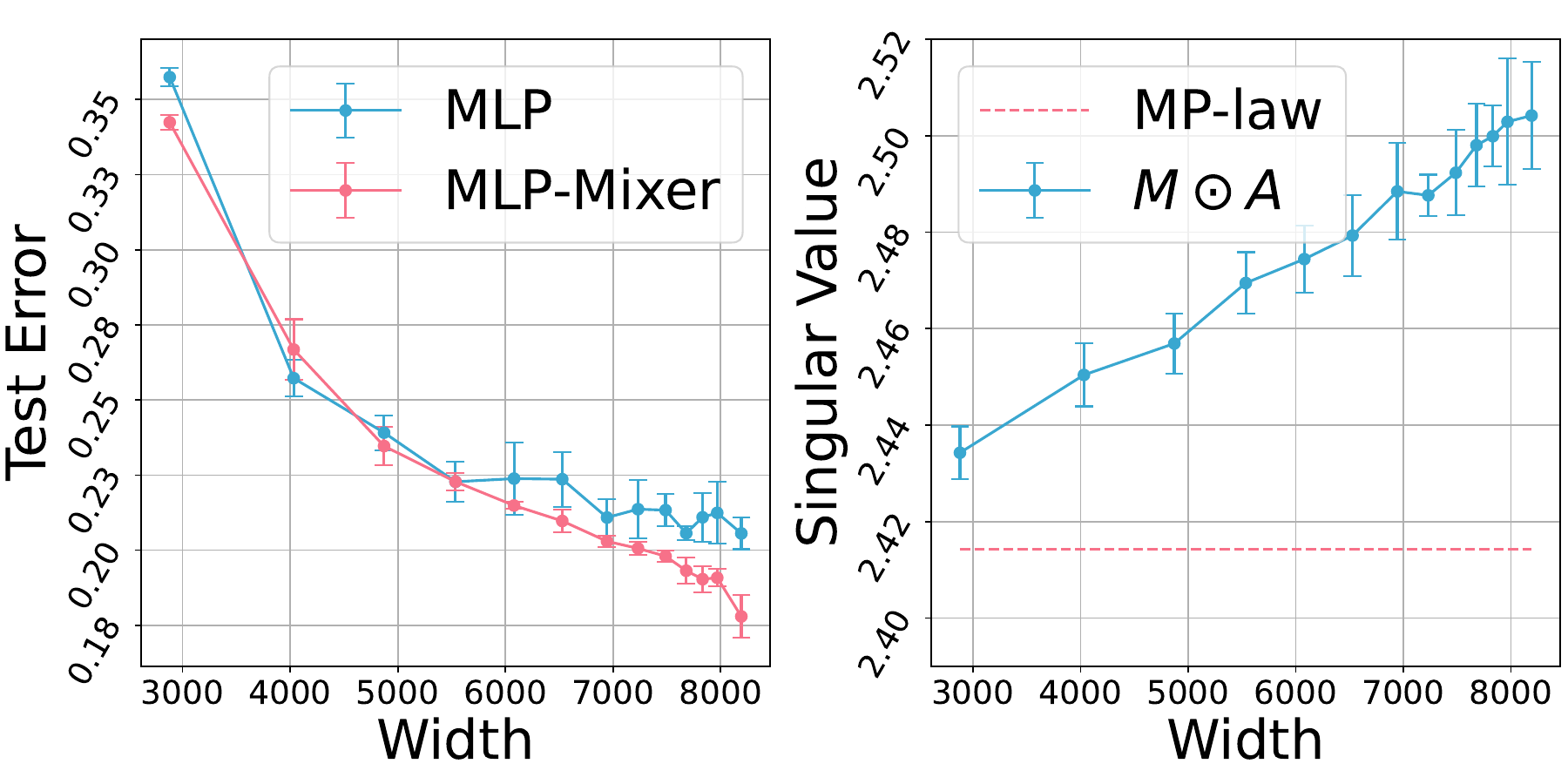}
    \caption{ (left) Test error of MLPs with sparse weights and MLP-Mixers with different widths $\gamma m$ under the fixed $\Omega$. We set $\Omega=2^{19}$, $S=C=(\Omega/\gamma)^{1/3}$, and $\gamma=2$. The x-axis represents the effective width $\gamma m$.
    (right) The blue line indicates the averaged singular values of the weight $M \odot A$ of SW-MLP over five trials with different random seeds. The red line indicates $c_\gamma$, which is the square root of the right edge of the MP-Law. }
    \label{fig:acc-spectrum}
\end{figure}

In the case of Mixers, each $\gamma C\times C$ weight $V$ (resp.\,$\gamma S \times S$ weight $W$) is initialized by i.i.d.\,random variables distributed with $\mathcal{N}(0,1/C)$ (resp.\,$\mathcal{N}(0,1/S))$. Then, by the theory of Marchenko-Pastur law~\cite{bai2010spectral}, the weight's maximal singular value is approximated by $c_\gamma:=1 + \sqrt{\gamma}$.  Since the Kronecker product with identity matrix does not change the maximal singular value, each maximal singular value of $V^\top \otimes I_S$ or $I_C \otimes W$ is approximated by  $c_\gamma$.

Consider the case of SW-MLP.  We initialize the entries of each mask matrix $M$ by i.i.d.\,Bernoulli random variables with the probability of being  $1/\sqrt{p}$ is $p$ and being $0$ is $1-p$. We initialize each weight $A$  by i.i.d.\,$\mathcal{N}(0,1/m)$.  Consider the  maximal singular value $\lambda_\mathrm{max}$ of $M \odot A$.
Set  $q= \sqrt{pm}$. Then by \citep[Theorem 2.9]{hwang2019local},   $\lambda_\mathrm{max}$ is approximated by $\sqrt{L_+}$ in the following sense:
\begin{align}
    | \lambda_\mathrm{max}^2 - L_+ |  \prec 1/q^4 + 1/m^{2/3},
\end{align}
where $\prec$ represents stochastic domination  \citep[Definition 2.3]{hwang2019local}, and 
\begin{align}\label{align:Lplus}
    L_+ = c_\gamma^2 + 3c_\gamma^2 \sqrt{\gamma} {(1-p) / q^2} + O({1 / q^4}).
\end{align}
Under the fixed $\Omega$, by (\ref{align:omega-p}), the dominant term of $L_+$ in (\ref{align:Lplus}) is linear in $m$ as follows:
\begin{align}
    { (1-p)/q^2} =  { \gamma m / \Omega} - { 1/ m} = O(m) \text{\ as \ } m \to \infty.
\end{align}
Therefore, the maximal singular value $\lambda_\mathrm{max}$ of $M \odot A$ increases as the width increases.
In \cref{sec:trainability}, we discuss the spectrum in large $\Omega$ limit and show the same tendency on the $\lambda_\mathrm{max}$ as large $m$ limit.

In \cref{fig:acc-spectrum}~(right), we observed the maximal singular value of $M\odot A$ increased by widening $m$ with fixed $\Omega$ and $\gamma$. In particular, the value was always higher than the theoretical value of Mixer's singular values. Such large singular values deteriorate the performance of SW-MLP.  Conversely,  we can enlarge the width and the sparsity of the MLP-Mixer without an undesirable increase in the maximal singular value.

\section{Beyond the naive MLP on the width}\label{sec:beyond}

As seen in Section~\ref{sec:acc_vs_sw}, MLP-Mixers have similar tendencies to SW-MLPs.
In much wider models, to continue comparing MLP-Mixer and unstructured sparse-weight MLP,  we need an alternative to static-masked MLP because of its huge computational costs, memory requirements (\cref{sec:cost}), and ill behavior on the spectrum (\cref{sec:spectrum}). 
Thus we further discuss partially destroying MLP-Mixer's structure and  propose an alternative model of SW-MLP, which is called random permuted mixer (RP-Mixer).
\begin{figure*}[t]
\centering

\includegraphics[width=0.9\textwidth]{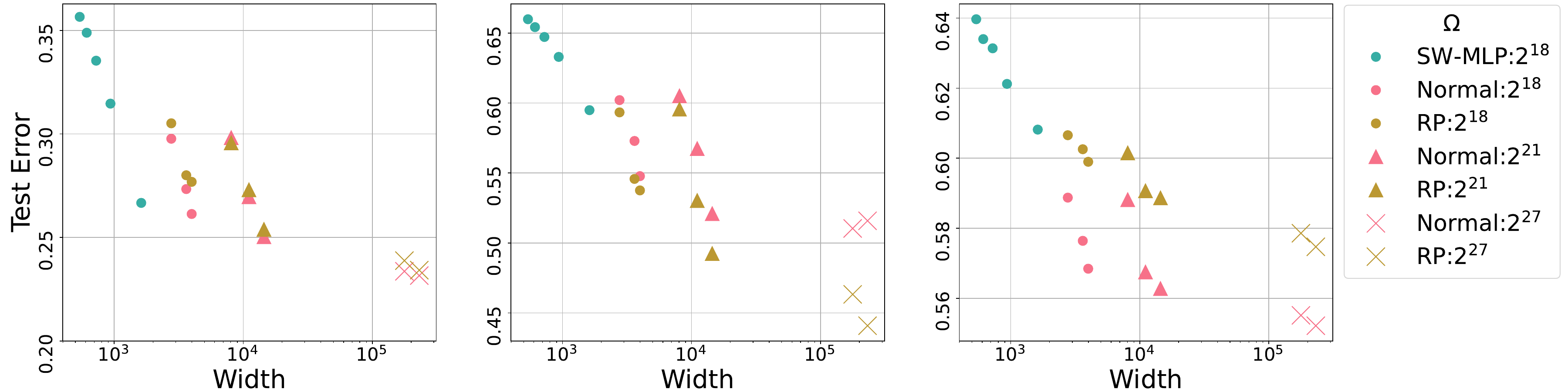}

\caption{Test error improved as the effective width increased. This figure presents S-Mixer, RP S-Mixer, and SW-MLP models on  CIFAR-10 (left), CIFAR-100 (center), and STL-10 (right). Experiments were conducted using three different random seeds, and the mean test error is depicted. The observed standard deviations were less than 0.026 for CIFAR-10, 0.056 for CIFAR-100, and 0.008 for STL-10.}
\label{fig:width}
\end{figure*}
\begin{table*}[t]
    \centering
    \begin{tabular}{c|| c | c  || c | c  }
       MLP & CIFAR-10 & CIFAR-100 &  max.\,width & $\# \text{connections} $  \\
       \hline
        $\beta$-LASSO \citep{neyshabur2020towards}&  14.81  &  40.44  &   - & 256M   \\
        Mixer-SS/8 &  15.91 ($\pm 1.55$) & 44.24($\pm 1.83$) &  $6.3 \times 10^4$ & 256M \\
        \bf{Mixer-SS-W} &  $\mathbf{12.07}$ ($\pm 0.47$)  &  $\mathbf{38.13}$($\pm 1.36$) & $\mathbf{1.2 \times 10^5}$ &  255M \\
        \hline
    \end{tabular}
    
    \begin{tabular}{c|| c || c | c | c | c }
       MLP & ImageNet-1k & max.\,width & $\Omega$  & S & C  \\
       \hline
        Mixer-B/16 \cite{Tolstikhin21mlp-mixer} &  23.56 &   $6.0 \times 10^5$ & $2.6 \times 10^8$   & 196 & 786  \\
        \bf{Mixer-B-W} &  $\mathbf{23.26}$ ($\pm$ 0.19 )  & $\mathbf{6.2\times 10^5}$ & $2.6 \times 10^8$  & 256 & 588 \\
        \hline
    \end{tabular}

    \caption{Test error on CIFAR-10/CIFAR-100/ImageNet-1k from scratch. (Upper Table) By setting $S$ and $C$ closer \textbf{under the same number of total connections} throughout layers, the maximal width of layers became larger in ours (Mixer-SS-W).  Its test error eventually improved more than $\beta$-LASSO. (Lower Table) By setting $S$ and $C$ closer \textbf{under the same $\Omega$}, the test error in ours (Mixer-B-W) improved than the original MLP-Mixer (Mixer-B/16).  Each experiment is done with three random seeds.
    }
    \label{tab:cifar}
\end{table*}

\subsection{PK family}
\label{sec:pk-family}
To introduce an alternative to sparse-weight MLPs, we propose a permuted Kronecker (PK) family as a generalization of the MLP-Mixer.

\paragraph{Permutation matrix:}
An $m \times m$ permutation matrix $J$ is a matrix given by $ (J x)_i = x_{\sigma(i)}  (i=1, 2, \dots, m)$
for an index permutation $\sigma$. 
In particular, the commutation matrix $J_c$ is a permutation matrix \citep{magnus2019matrix}.
For any permutation matrix $J$, $x \in \R^m$, $J\phi(x)=\phi(Jx)$.

\begin{defi}[PK layer and PK family]
\label{def:pk}
   Let $J_1, J_2$ be $m \times m$  permutation matrices. For $X \in \mathbb{R}^{n_1\times n_2}$, we define the PK layer as follows:
   \begin{align}
 \text{PK-Layer}_W(X;{J_1,J_2}) :=   \phi [  J_2 (I_{n_1} \otimes W ) J_1 \text{vec}(X) ], \nonumber
\end{align}
where we set $m=n_1 n_2$, $W \in \mathbb{R}^{n_2 \times n_2}$. We refer to the set of architectures whose hidden layers are composed of PK layers as the PK family.
\end{defi}
Since $J_c$ is a permutation matrix, the normal S-Mixer and MLP-Mixer belong to the PK family (See Section~B.3 for the details). The important point of the PK-Layer is that its width $m$ is possibly large, but there is {\it no need to explicitly preserve the $m\times m$ weight matrix} in memory. We can compute the forward signal propagation by a relatively small matrix multiplication in the same manner as the MLP-Mixer:
First, $J_1 \text{vec}(X)=:y$ is a rearrangement of $X$ entries. Next, we compute pre-activation by using the matrix product $(I_{n_1}\otimes W)y= W\text{mat}(y)$. Finally, we apply entry-wise activation and rearrangement by $J_2$. Thus, the PK layer is memory-friendly, whereas the naive dense MLP requires preserving an $m\times m$ weight matrix and is computationally demanding.

\subsection{Random Permuted Mixers}\label{sec:rp-mixer}
In normal Mixers, $J_1$ and $J_2$ are restricted to the identity or commutation. This means that the sparse weight matrices of the effective MLP are highly structured because their block matrices are diagonal. 
To destroy the structure, we introduce RP-Mixers. A \emph{RP S-Mixer} has ($J_1$,$J_2$) in each PK layer, which is given by random permutation matrices as $U=\text{PK-Layer}_W(X;{J_1,J_2})$ and $\text{PK-Layer}_{V^\top}(U;{J_1',J_2'})$.
Similarly, for a \emph{RP MLP-Mixer}, we set the PK layer corresponding to token mixing and channel mixing to the random permutation matrices.
From the definition of the PK layer (\ref{def:pk}), this is equivalent to the effective MLP with width $SC$ (and $\gamma SC$ for the MLP-Mixer) and sparse weight
\begin{equation}
  W_{eff}=  J_2 (I_{n_1} \otimes W ) J_1.
  \label{Weff}
\end{equation}
Because $(J_1,J_2)$ are random permutations, the non-zero entries of $W_{eff}$ are scattered throughout the matrix. In this sense, RP Mixers seemingly become much closer to random sparse weights than the normal Mixers. \cref{fig:schematic}(d) illustrates this scattered random weight configuration, while \cref{fig:mat_visual} presents an actual numerical example.
Since the permutation matrices are orthogonal matrices,  $W_{eff}$'s singular values remain identical to those of the normal Mixer, thereby preserving the spectrum discussed in \cref{sec:spectrum}.

\label{sec:experiments}

\subsection{Increasing width}
\label{sec:experiments-2}

\begin{figure*}[h]
\centering
\includegraphics[width=0.83\textwidth]{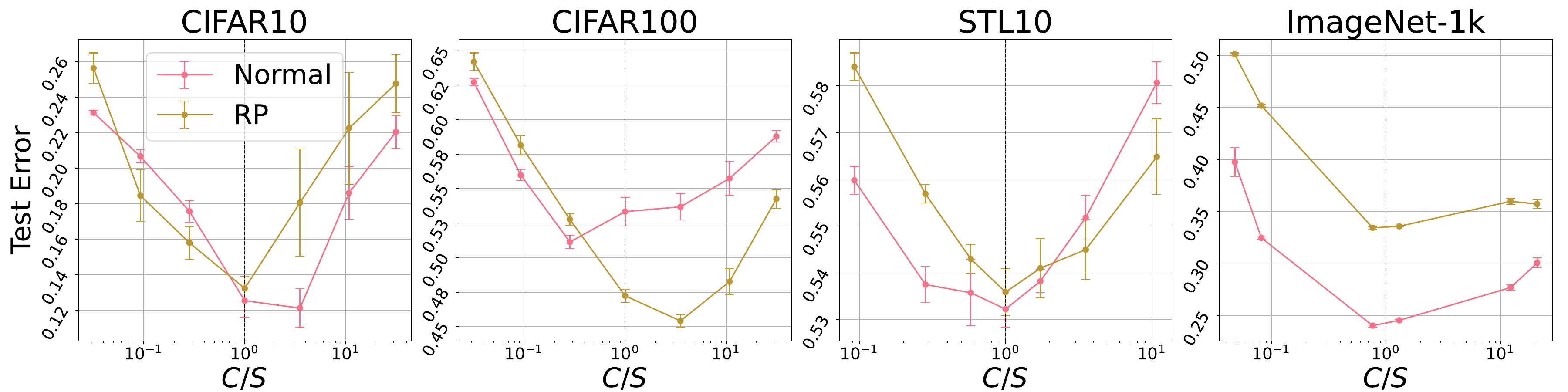}

\caption{ PK family achieves the lowest test error around $C=S$.   S-Mixers on (a)  CIFAR-10, (b) CIFAR-100, (c) STL-10. MLP-Mixers on (d) ImageNet-1k. 
Red line: Normal Mixers, yellow line: RP Mixers, dashed lines:  $C=S$.  
}
\label{fig:SC}
\end{figure*}

\begin{figure*}[t]
\centering
\includegraphics[width=0.832\textwidth]{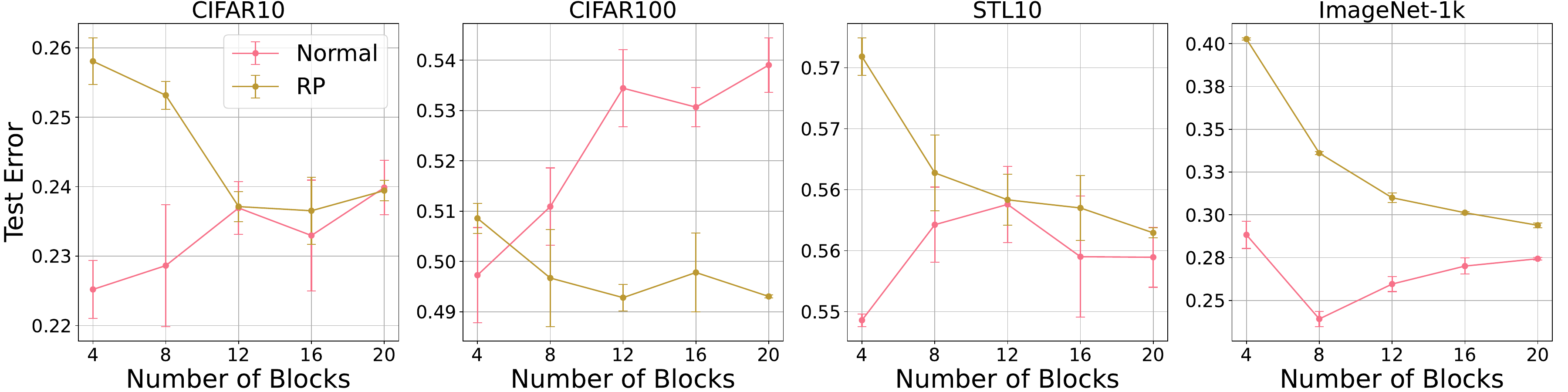}

\caption{RP Mixers can become comparable to or even beat normal ones if the depth increases.  We set $C=S=128$.}
\label{fig:depth}
\end{figure*}

\cref{fig:width} shows that the test error improves as the effective width of the Mixers increases. We trained the normal and RP S-Mixers for various values of $S$ and $C$ with fixed $\Omega$.
The normal and RP S-Mixers show similar tendencies of increasing test error with respect to the effective width $m$. The normal and RP MLP-Mixers also show similar tendencies as is shown in \cref{sec:increasing_width_mlpmixer_appendix}.
Because the static-mask SW-MLP requires an $m \times m$ weight matrix,  an SW-MLP with a large $\Omega$ cannot be shown in the figure. In contrast, we can use a large effective width for the Mixers, and the error continues to increase as the width increases. This can be interpreted as the Mixers realizing a width setting where naive MLP cannot reach sufficiently. Eventually, the figure suggests that such an extremely large width is one of the factors contributing to the success of the Mixer.

\cref{tab:cifar} shows a comparison of MLP-Mixer with a dynamic sparsity $\beta$-LASSO~\citep{neyshabur2020towards}.  We found a wider Mixer (Mixer-SS-W) has better performance than $\beta$-LASSO. \cref{tab:cifar} shows a comparison of Mixer-B/16~\citep{Tolstikhin21mlp-mixer} and a wider Mixer (Mixer-B-W). By setting $S$ and $C$ closer under fixed $\Omega$, the maximal width of layers became larger in ours (Mixer-B-W). Its test error eventually improved more than the original MLP-Mixer (Mixer-B/16).
In both results, the wideness improved the performance even if $\Omega$ is fixed.

\subsection{Performance at the optimal width and sparsity}
\label{sec:experiments-1}

 \cref{fig:SC} confirms that the maximum width (\ref{max_width}) derived from Hypothesis~\ref{hyp}  adequately explains the empirical performance of the Mixers.  Models were trained using supervised classifications for each dataset.  
For CIFAR-10, CIFAR-100 and STL-10, we trained normal and RP S-Mixers.
We fixed the dimension of the per-patch FC layer and changed $S$ and $C$ while maintaining a fixed number of $\Omega$. 
It can be observed that the test error was minimized around $C=S$, as expected from Hypothesis~\ref{hyp} and  (\ref{max_width}). This tendency was common to the normal and RP Mixers.  
For ImageNet-1k, we trained normal and RP MLP-Mixers. Similarly, its performance is maximized around $C=S$. 
We observed a similar tendency also in different settings of hyparparamters in \cref{sec:experiments-3} and \cref{sec:replacement}.

\noindent
{\bf Remark on depth:} 
As is shown in \cref{fig:depth}, we observed that RP Mixers tended to underperform at limited depths but could achieve comparable or better results than normal Mixers with increased depths. This seems rational due to deep RP-Mixer's resistance to overfitting or shallow RP-Mixer's small receptive fields (see \cref{sec:depth} for the details). We also confirmed that this dependence on depth did not change the fundamental similarity regarding the width and sparsity.

\section{Conclusion and future directions} 
This work provides novel insight that the MLP-Mixer effectively behaves as a wide MLP with sparse weights. 
The analysis in the linear activation case elucidates the implicit sparse regularization through the Kronecker-product expression and reveals a  connection to Monarch matrices. The SW-MLP, normal and RP Mixers exhibit a quantitative similarity in performance trends,  verifying that the sparsity is the key mechanics underlying the MLP-Mixer. Maximizing the effective width and sparsity leads to improved performance. We expect that the current work will serve as a foundation for exploring further sophisticated designs of MLP-based architectures
and the efficient implementation of neural networks with desirable implicit biases.

Exploring the potential of MLPs with structured weights further will be interesting. Evaluating the optimal width and sparsity theoretically could be an exciting research topic \cite{edelman2023pareto}. As we noted, the solvability of global minima and dynamics in mixing layers, even with linear activation, remains uncertain, and the theory has yet to fully address or circumvent this issue. It would be also interesting to clarify whether other potential candidates for memory-friendly architectures with desirable inductive biases. In particular, weight sharing is known to perform well for image domains, occasionally outperforming networks without weight sharing \cite{ott2020learning}. Given that the Mixers can be seen as approximations of MLPs with weight-shared Monarch matrices,  
 it will be an interesting theme to evaluate the validity of such  approximations.

\bibliographystyle{icml2024}
\bibliography{arxiv_ref}

\newpage
\appendix
\onecolumn

\setcounter{section}{0}
\renewcommand{\thesection}{\Alph{section}}
\renewcommand{\theequation}{S.\arabic{equation}}
\renewcommand{\thefigure}{S.\arabic{figure} }
\renewcommand{\thetable}{S.\arabic{table} }

\setcounter{equation}{0}
\setcounter{figure}{0}
\setcounter{table}{0}

\noindent
\arxiv{
}

\section{Details of Architectures}\label{sec:arch}

Here, we overview more technical details of all models: MLP-Mixer, Simple Mixer (S-Mixer), and MLP with sparse weights (SW-MLP). 
In Section \ref{secA-1}, we introduce the transformation from the input image to the first hidden layer. In Section \ref{secA-2}, we overview some detailed formulation of the models including skip connection and layer normalization. 


\begin{table}[bh]
    \centering
    \caption{Summary table of models.  The notation hidden in the block means whether the model has a hidden layer in each MLP block. The symbol (*) indicates the setting depends on the kind of experiment.  Here, GAP is the global average pooling. }
    \begin{tabular}{cccccc}
    \toprule
       Model           &  Per-patch FC & Skip-conn. & Layer-norm  & Hidden in the block & GAP\\
       \hline
      SW-MLP           & \checkmark    & \checkmark      & \checkmark  & (*)             & \checkmark\\
      (RP) MLP-Mixer   &  \checkmark   & \checkmark      & \checkmark  & \checkmark      & \checkmark \\
      (RP) S-Mixer     & \checkmark    & \checkmark      & \checkmark  &                 & \checkmark \\
      Shallow Kronecker &  & & & \\ 
      Shallow Monarch & &  & &  \\
      \hline
    \end{tabular}
    \label{tab:models}
\end{table}

\subsection{Per-patch FC Layer}
\label{secA-1}
The first layer of the MLP-Mixer is given by the so-called per-patch FC layer, which is a single-layer channel mixing.
In all experiments, for a  patch size $P$, the input image is decomposed into $HW/P^2$ non-overlapping image patches with size $P \times P$;  we rearrange the  $H \times W$ input images with $3$ channels into a matrix whose size is given by $(HW/P^2) \times 3P^2 = S_0 \times C_0$. 
For the rearranged image $X \in \mathbb{R}^{S_0\times C_0}$, the per-patch fully connected (FC) layer is given by
\begin{equation}
   Y =  X W^\top,
\end{equation}
where $W$ is a $C \times C_0$ weight matrix.
We use the per-patch FC layer not only for Mixers but also for SW-MLP.

\paragraph{Remark on per-patch FC layer:}
The original study set the size of the mixing layers to $S=S_0$. 
In contrast, to investigate the contribution of each input image size and hidden layer size independently, it is rational to change $(S,C)$ independent of $(S_0, C_0)$. Therefore, we make the per-patch FC transform the input size $C_0$ to the output size $C$ and the first token mixing layer transform $S_0$ to $S$.

\subsection{MLP-Mixer and S-Mixer}
\label{secA-2}

Let us denote a block of the MLP-Mixer by
\begin{align}
    f_{W_1, W_2}(X) = \phi(XW_1^\top) W_2^\top,
\end{align}
and that of the S-Mixer by
\begin{align}
    f_{W_1}(X) = \phi(X W_1^\top ).
\end{align}
We set $\phi = \mathrm{GELU}$.

\subsubsection{MLP-Mixer}

We set the layer normalization ($\LN$)  by 
\begin{align}
    \LN(X)=   { X - m(X) \over \sqrt{v(X) + \epsilon} } \odot \gamma + \beta,  \quad X \in \R^{S \times C}, 
\end{align}
where $\odot$ denotes the Hadamard product , $m(X)$ (resp.\,$v(X)$) is the empirical mean (resp.\,the empirical variance) of $X$ with respect to  the channel axis, and $\gamma, \beta$ are trainable parameters. We set $\epsilon = 10^{-5}$ in all experiments.

In the implementation of fully connected layers, we use only the right matrix multiplications in the same way as the original MLP-Mixer \cite{Tolstikhin21mlp-mixer}. A token-mixing block $X \mapsto U$ of MLP-Mixer is given by 
\begin{align}
    U = X  +  f_{W_1, W_2}(  \LN(X)^\top )^\top,
\end{align}
where $W_1$ is $S \times \gamma S$ and $W_2$ is $\gamma S \times S$.
Similarly, we set a channel-mixing block  $U \mapsto Y$ as
\begin{align}
    Y = U + f_{W_3, W_4}(\LN(U)),
\end{align}
where $W_3, W_4$ are weight matrices.

We refer to the composed function $X \mapsto Y$ of the token-mixing block and the channel-mixing one as \emph{a base block} of the MLP-Mixer. 
The MLP-Mixer with $L$-blocks is composed in the order of the per-patch FC layer, the $L$ base blocks, and the global average pooling with the layer normalization,  and the last fully connected classification layer.

\subsubsection{S-Mixer}
The S-Mixer without random permutations is implemented by replacing the MLP-block $f_{W_1, W_2}$ and $f_{W_3, W_4}$ in the MLP-Mixer with  FC blocks. That is,   token-mixing and channel-mixing blocks are given by 
\begin{align}
    U &= X +  f_{W}(\LN(X)^\top)^\top, \\
    Y &= U + f_{V}(\LN(U)),
\end{align}
where $W$ and $V$ are weight matrices.
The transpose of the input matrix in the token-mixing block is implemented by rearrangement of entries. 
We decided to apply both skip-connection and layer normalization even in the S-Mixer. This is a rather technical requirement for ensuring the decrease of training loss in deep architectures.

\subsubsection{Mixers with Permuted Kronecker Layers}
Here we implement generalized MLP-Mixer and S-Mixer with permutation matrices and PK-layers.
Recall that for any matrix $X$,
\begin{align}
    X^\top =  \mathrm{Mat} (  J_c  \mathrm{vec} (X) ), 
\end{align}
where $J_c$ is the $m \times m$ commutation matrix.
Therefore, the token-mixing block of the S-Mixer is 
\begin{align}
    U &= X +   \mathrm{Mat} \circ \ J_c^\top \circ \mathrm{vec} \circ f_{W} \circ \mathrm{Mat}  \circ J_c \circ \mathrm{vec} \circ \LN(X)\\   
    & = X +  \mathrm{Mat} \circ \text{PK-Layer}_W (\LN(X); J_c, J_c^\top).
\end{align}
Similarly, the channel-mixing block of the S-Mixer is equal to 
\begin{align}
    Y = U +  \mathrm{Mat} \circ \text{PK-Layer}_V (\LN(U); I, I),
\end{align}
where $I$ is the identity matrix.
Note that skip-connections  gather $J_c$ and $J_c^\top$ in the same mixing block for compatibility in shapes of hidden units. 

To get examples of PK family and to generalize Mixers, we implement the random permuted (RP) S-Mixer with skip-connections by replacing $J_c$ and $J_c^\top$ with i.i.d.\,random permutation matrices $J_1$ and $J_2$: 
\begin{align}
    U = X +  \mathrm{Mat} \circ \text{PK-Layer}_W (\LN(X); J_1, J_2).
\end{align}
We implement the random permutations by random shuffling of output $m=SC$ indexes of vectors. We freeze it during the training step.  Note that we avoid using an $m \times m$ matrix representation of $J_x$ for memory efficiency.  We implement the random permuted (RP) MLP-Mixer by  the same way as the RP-S-Mixer.

\subsubsection{The skip-connection in the first block}

The first token-mixing block has the input shape $(S_0, C)$ and the output shape $(S, C)$.
However, we need to change $S$ with fixing $S_0$ in some experiments.
To control the difference of $S_0$ and $S$,  we set the first token-mixing block as follows: 
\begin{align}
     U =  \text{SkipLayer}(X) + \text{PK-Layer}_W(\text{LN}(X); J_1, J_2 ) ,
\end{align}
where the skip layer is given by 
\begin{align}
    \text{SkipLayer} (X) =  \LN( \Tilde{W} X ),
\end{align}
where $\Tilde{W}$ is a $S \times S_0$ weight matrix.
For a fair comparison, we use the skip layer even if $S = S_0$ in the experiments that we sweep $S$.  We use the same setting for the MLP-Mixer as for the S-Mixer. 

\subsubsection{Sparse-Weight (SW) MLP}
Let $ 0< p \leq 1$ and $m \in \N$.
We implement  each matrix  of  a static sparse weight FC block with the freezing rate $p$ as follows: 
\begin{align}\label{align:sw-mlp}
 x \mapsto    x +   \phi( ( M \odot W ) \LN(x)), \quad x \in \R^m,
\end{align}
where $M$ is the mask matrix whose $m^2 p$ entries are randomly chosen and set to be one with a probability $p$  and the others are set to be zero with a probability $1-p$.
The mask matrix $M$ is initialized before training and it is frozen during training.

We also consider the SW-MLP consists of sparse weight MLP-blocks as follows:
\begin{align}\label{align:sw-mlp-hidden}
 x \mapsto    x +   \phi \left( ( M_2 \odot W_2 ) \phi( ( M_1 \odot W_2 ) \LN(x)) \right), \quad x \in \R^m.
\end{align}    
$W_1$ and $W_2$ are weight matrices with hidden features $\gamma m$, where $\gamma$ is an expansion factor. $M_1, M_2$ are mask matrices whose $\gamma m^2 p$ entries are randomly chosen and set to be one with a probability $p$ and the others are set to be zero with a probability $1-p$.

 SW-MLP with $L$-blocks is composed in the order of the per-patch FC layer,  vectorization, $L$ static sparse weight FC blocks (or MLP blocks), and the last classification FC layer.

\section{Analysis}

\subsection{Derivation of Proposition 4.1}\label{sec:proof_expression}

For $H=\phi(\phi(WX)V)$, by using $\text{vec}(W X V) =  (V^\top \otimes W) \text{vec}(X)$,  we have 
\begin{align}
    \text{vec}(H) &= \phi  ((V^\top \otimes I_S)   \text{vec}(\phi(WX)) \\ 
     &= \phi ((V^\top \otimes I_S)   \phi ((I_C \otimes W) x)). 
\end{align}
Because $J_c^\top (A \otimes B) J_c = (B \otimes A)$ \cite{magnus2019matrix} and any permutation matrix $J$ is commutative with the entry-wise activation function: $J \phi (x)  = \phi( Jx)$, we obtain
\begin{equation}
 \text{vec}(H) =   \phi (J_c^\top(  I_S \otimes V^\top)   \phi (J_c(I_C \otimes W) x)).
\end{equation}

The skip-less MLP-Mixer (\ref{eq:mixer}) is expressed as follows:
$Y=\text{Channel-MLP}(\text{Token-MLP}(X)))$, and then 
\begin{align*}
u&= \phi (J_c (I_C \otimes W_2) \phi((I_C \otimes W_1) x)),\\
y&= \phi(J_c^\top (  I_S \otimes W^\top_4 )  \phi((  I_S \otimes W^\top_3 ) u) ). \label{eq:effMLPmixer}    
\end{align*}
where $u=\text{vec}(U)$ and $y=\text{vec}(Y)$.

It may be informative that a similar transformation between the matrix and vector is used in a completely different context of deep learning, that is, the Kronecker-factored Approximate Curvature (K-FAC) computation for natural gradient descent \cite{martens2015optimizing}. K-FAC assumes layer-wise 
preconditioner given by the Kronecker product, that is, $(B \otimes A)^{-1}\text{vec}(\nabla_W Loss(W))$ where $A$ and $B$ correspond to the Gram matrices of the forward and backward signals. This K-FAC gradient can be computed efficiently because it is reduced to a matrix computation of $A^{-1} \nabla_W Loss(W) (B^\top)^{-1}$. Therefore, the trick of formulating a large matrix-vector product for the product among relatively small matrices is common between K-FAC and the aforementioned effective expression.

\subsection{Product of Random Permutations}
The uniformly distributed random  $m \times m$ permutation matrix is given by  $J=J_g$ where $g$ is the uniformly distributed random variable on the permutation group $S_m$ of $m$ elements. 
Then the uniform distribution over $S_m$ is the Haar probability measure, which is translation invariant (see \cite{folland2013real} for the detail), that is, $J =J_{\sigma}J_{g}$ is also uniformly distributed if $\sigma$ is $S_m$-valued uniformly distributed random variables and $g \in S_m$ is constant.  Therefore, $J= J_{\sigma} J_{\rho}$ is a uniformly distributed random permutation matrix for independent and uniformly distributed random variables $\sigma$ and $\rho$ on $S_m$. 

\subsection{Representation as a PK-family}
 By  (\ref{eq11:0502}), one can see that   
 the block of the S-Mixer is 
 \begin{align}
  U=\text{PK-Layer}_W(X;{I,J_c}),\    \text{PK-Layer}_{V^\top}(U;{I,J_c^\top})
 \end{align}
 
For MLP-Mixer, the
Token-MLP is 
\begin{align}
 U= \text{PK-Layer}_{W_2}( \text{PK-Layer}_{W_1}(X;{I,J_1});J_1^\top,J_c)   
\end{align}
and Channel-MLP is 
\begin{align}
 \text{PK-Layer}_{W_4^\top}( \text{PK-Layer}_{W_3^\top}(U;{I,J_2});{J_2^\top,J_c^\top})   
\end{align}
for the arbitrary permutation matrices $J_1$ and $J_2$.

\subsection{Proof of implicit regularization}\label{sec:proof_regularization}
Here we prove \cref{prop:regularization} and show a connection between that and \cite{hoff2017lasso}.
Since each weight $V, W$ are $C^2, S^2$ dimensional vectors, we only need to show the following \cref{lem:regularization_vec}.
 
\begin{lem}\label{lem:regularization_vec}
Let $m,n \in \N$. We write $||x||_p$ for the $L_p$ norm of any real vector $x$.
Let $f: \R^{mn} \to \R$. Set $h: \R^{mn} \to \R$, $g: \R^m \times \R^n \to \R$ as follows:
\begin{align}
    \phi(\beta) &= f(\beta) + \lambda ||\beta||_2,\\
    h (\beta) &=  f(\beta) +  { \lambda \over \sqrt{mn}}||\beta||_1,\\
    g (u,v) &=  f( u \otimes v) + \lambda (|| u ||_2^2 + || v||_2^2)/2.
\end{align}
Set
\begin{align}
    \Theta_p := \{ u \otimes v \mid u \in \R^m , v \in \R^n\}.
\end{align}
Then it holds that
   \begin{align}\label{align:regularizatoin_vec}
       \inf_{u \in \R^m,v \in \R^n} g(u,v) = \inf_{\beta \in \Theta_p} \phi(\beta)  \geq \inf_{\beta \in \R^{mn}} h(\beta).
   \end{align}
\end{lem}

\begin{proof}
Since  $||u \otimes v||_2 = ||u||_2 || v||_2$, we have
\begin{align}
       \inf_{u \in \R^m,v \in \R^n} g(u,v) = \inf_{\beta \in \Theta_p} \left[ f( \beta) + \lambda/2 \inf_{v \in \R^n  } (|| \beta ||_2^2/||v||_2^2 + || v||_2^2) \right].
\end{align}
The inner $\inf$ achieves the maximum when $||v||_2^2 = ||\beta||_2$, and the maximum is $||\beta||_2$. (Note that it is {\bf{not squared}} L2-norm.)
Therefore, 
\begin{align}
      \inf_{u \in \R^m,v \in \R^n} g(u,v) = \inf_{\beta \in \Theta_p }(f(\beta) + \lambda ||\beta||_2).
\end{align}
Lastly, by the H\"{o}lder's inequality, it hold that $||\beta||_1 \leq  ||\beta||_2 \sqrt{mn}$, which complete the proof.
\end{proof}

\newcommand{\onevec}{\mathbf{1}}
To describe the connection of the implicit L1 regularization in the case of  Hadamard product \cite{hoff2017lasso}, consider matrix $\mathbf{1}\mathbf{1}^\top$, where $\mathbf{1}$ is the
vector with all entries are one.  For simplicity, consider the case $S=C$. It is easy to recover the general case by changing the proportional coefficients.
Then we have
\begin{align}
\Tr ( W^\top W \otimes \onevec \onevec^\top) + \Tr ( \onevec \onevec^\top \otimes V^\top V)  =  C^2  ||W||_F^2 + S^2 || V||_F^2 = C^2 (|| W||_F^2 + ||V||_F^2)
\end{align}
Put 
\begin{equation}
g(W,V):=\mathcal{L}(W \otimes V) + \lambda/2 (\|W\|^2_F + \|V\|^2_F).
\end{equation}
We have 
\begin{equation}
g(W,V) = \mathcal{L}(W \otimes V) + \frac{\lambda}{2C^2}( \Tr ( W^\top W \otimes \onevec \onevec^\top) + \Tr ( \onevec \onevec^\top \otimes V^\top V))
\end{equation}
Here consider the vectorization: $u(W):=\mathrm{vec}(W \otimes \onevec \onevec^\top)$, $v(V):=\mathrm{vec}(\onevec \onevec^\top \otimes V)$. 
Then 
\begin{align}
    u (W) \odot v(V) = \mathrm{vec}(W \otimes V).
\end{align}
Thus we have
\begin{align}
\inf_{W,V} g(u,v) &=  \inf_{W,V} \mathcal{L}(u \circ v) +  \frac{\lambda}{2C^2} (\|u\|^2_2+\|v\|^2_2),
\end{align}
Because of the domain of $u$ and $v$, we have
\begin{align}
\inf_{W,V} g(u,v) &\geq   \inf_{u, v \in \R^{S^2C^2}} \mathcal{L}(u \circ v) +  \frac{\lambda}{2C^2} (\|u\|^2_2+\|v\|^2_2).
\end{align}
By \citet{hoff2017lasso}, the right-hand side is equal to 
\begin{align}\label{align:reguralization_factor}
\inf_{\beta \in \R^{S^2 C^2}} \mathcal{L}(\beta) +  \frac{\lambda}{2C^2} \|\beta\|_1.
\end{align}
Thus the inequality for the Kronecker product has a relationship with \cite{hoff2017lasso} from the vectorization and Kronecker product with $\mathbf{1}\mathbf{1}^\top$.

Let us focus on the normalization factor $1/mn$ and $1/C^2$.
In general, there is a trivial relationship between L1 and L2 regularization as follows:
For $\theta \in \R^M$,
\begin{equation}
\inf_{\theta} \mathcal{L}(\theta) +  \sqrt{M} {\lambda} \|\theta\|_2 \geq \inf_{\theta} \mathcal{L}(\theta) +  {\lambda} \|\theta\|_1,
\end{equation}
this directly follows from the H\"{o}lder's inequality.
However, in our situation, the parameter space of $W \otimes V$ is a subspace of $\R^{S^2C^2}$, so it is not clear whether the same constant regularization for both L1 and L2 regularization is suitable. 
At the initialization, we set entries of $W, V$ independently distributed with $\mathcal{N}(0,1/C)$, thus by the law of large numbers, 
\begin{equation}
\|W\|_F^2, \|V\|_F^2 = O(C).
\end{equation}
In the case of dense MLP, we initialize $\beta$ as $\mathcal{N}(0,1/SC)$, thus
\begin{equation}
  \|\beta\|_1 = O(S^2C^2/\sqrt{SC}) = O(C^3).
\end{equation}
 Thus
\begin{align}
{      \|\beta\|_1 \over \| W\|_F^2}, {\|\beta\|_1 \over \| V\|_F^2 } = O(C^2).
\end{align}
Therefore, the normalizing factor  $1/C^2$ in (\ref{align:reguralization_factor}) matches the scale transformation according to the dimension of parameter spaces. Without the normalizing factor, the L1 regularization term will be $C^2$ times larger than the loss $\mathcal{L}$.

\section{Experimental Setting}\label{sec:experimental_setting}

\subsection{Figure \ref{fig:cka_fr_acc}}

We set the number of blocks of the MLP-Mixer, denoted by \(L\), to 3. Both the Token-mixing block and the Channel-Mixing block are present \(L\) times each, resulting in a total of 6 blocks when counted separately. We also set \(\gamma=2\).

For the comparison, the sparse-weight MLP replaces the components within the MLP-Mixer, namely \((I_C \otimes  W_1), (I_C \otimes W_2)\) and \(( W_3^\top \otimes  I_S), (W_4^\top \otimes I_S)\), with the form \(M \odot A\). In this context, there's no distinction between the token-mixing and the channel-mixing blocks, leading to a total of 6 blocks.

In \cref{fig:cka_fr_acc} (a), we compare the MLP-Mixer with \(S=C=64, 32\) and the SW-MLP. The sparsity of the SW-MLP is taken as \(2^{-n}\) where \(n\) ranges from 0 to 10.  We set the patch size as 4. Each network is trained on CIFAR10 with a batch size of 128, for 600 epochs, a learning rate of 0.01, using auto-augmentation, AdamW optimizer, momentum set to 0.9, and cosine annealing. We utilize three different random seeds for each training.

\cref{fig:cka_fr_acc}(b) shows the CKA (for a specific seed) of the MLP-Mixer with \(S=C=64\) and the MLP with a sparsity of 1/64, based on the results from \cref{fig:cka_fr_acc}(a). However, the features targeted for CKA are taken from the layer just before the skip-connection in each block, and they are displayed on the axis in the order of proximity to the input, labeled as 1 through 6. \cref{fig:cka_fr_acc}(c) similarly compares the dense MLP (i.e.\,sparsity$=1$).

\cref{fig:cka_fr_acc}(d) shows the test error of the shallow MLP with monarch matrix weight and  Kronecker weight matrix. The training uses MNIST, with optimizer adamw with 200 epochs, and the cosine annealing of leaning rate with an initial value of 0.01. 
\cref{tab:models} summarizes the difference of the shallow models between other treated models.

\subsection{Figure~\ref{fig:acc-spectrum}}
In \cref{fig:acc-spectrum}~(left), we compare the MLP-Mixer and the MLP under the same training settings as in \cref{fig:cka_fr_acc}(a). Now, here \(\Omega=2^{19}\) and $\gamma=2$, with \(C\) values being 4,8,12, 16, 20, 24, 28, 32, 36, 40, 44, 48, 64. The \(S\) values are determined corresponding to \(C\) while keeping \(\Omega\) fixed, resulting in values: 360, 252, 203, 173, 152, 136, 124, 113, 104, 96, 89, 83, 64. We utilized four different random seeds for each training.

In \cref{fig:acc-spectrum}~(right),
we used the same values on $\Omega, \gamma$ and $C,S$ as the \cref{fig:acc-spectrum}~(left). We run 10 trials with different random seeds to plot the singular values of the sparse weight matrix.

\subsection{ \cref{tab:cifar}}

\paragraph{CIFAR10, CIFAR100} We used single Tesla V100 GPU to compute the runtime. The training is by AdamW with the mini-batch size 128. We used a 32-bit float. The runtime is averaged over 600 epochs.
The averaged runtime is on four different random seeds.

For our experiments, we utilized Tesla V100 GPUs, accumulating approximately 300 GPU hours. The networks were trained on datasets, either CIFAR-10 or CIFAR-100. 
We set \(L=2\) and \(\gamma=4\) for the Mixer-SS/8 and Mixer-SS-W.
For the Mixer-SS/8 configuration, the parameters were set as \(p=8\), \(S=16\), and \(C=986\). In our wider setting (Mixer-SS-W), the parameters were defined as \(p=4\), \(S=64\), and \(C=487\). 
The training was conducted over 4000 epochs with the cosine-annealing. We employed the AdamW optimizer and incorporated auto-augmentation techniques. The chosen mini-batch size was 1024, and experiments were run with three distinct random seeds. To optimize results, we conducted a hyper-parameter search for the best initial learning rate from the set \(\{ 0.04, 0.05, 0.06\}\). The rate that provided the highest accuracy, when averaged over the random seeds, was subsequently chosen for experiments. Specifically, the learning rate 
0.06 was selected for Mixer-SS/8 on CIFAR-100, while 
0.04 was the preferred rate for all other scenarios.

\paragraph{ImageNet-1k}
For the Mixer-B-W, we set $L=8,p=14, C=588,S=256$, and $\gamma=4.57$. The other settings are the same as the other experiments on ImgeNet-1k (\cref{ssec:fig:sc:param}).

\subsection{ Figure \ref{fig:width} }\label{ssec:fig2}

We utilized Tesla V100 GPUs and approximately 400 GPU hours for this experiment. We trained three types of MLPs; S-Mixer, RP S-Mixer, and SW-MLP architectures.
All MLPs incorporated a per-patch FC layer as the first block, with a patch size of $P=4$. The input token size was fixed at $S_0=(32/P)^2=64$.
We trained the models on the CIFAR-10, CIFAR-100, and STL-10 datasets, along with data augmentations such as random cropping and random horizontal flipping. The input images were resized to a size of $32 \times 32 \times 3$. We employed Nesterov SGD with a mini-batch size 128 and a momentum of 0.9 for training, running for 200 epochs. The initial learning rate was set to 0.02, and we used cosine annealing for learning rate scheduling. To ensure robustness, we conducted three trials for each configuration and reported the mean and standard deviation of the results. Unless otherwise specified, these settings were used throughout the whole study on CIFAR-10, CIFAR-100, and STL-10.

\paragraph{(i) S-Mixer and RP S-Mixer} 
We conducted training experiments on the S-Mixer architecture with eight blocks. In order to explore various cases of integer pairs $(S,C)$ that approximately satisfy the equation
\begin{align}
\Omega = \frac{CS^2 + SC^2}{2}. \label{align:omega-non-gamma}
\end{align}
The number of connections, denoted as $\Omega$, was fixed at $\Omega = 2^{18}, 2^{21}$ and $2^{27}$.
For each value of $\Omega$, the pairs $(C,S)$ were chosen in a symmetric manner. It should be noted that if $(C,S) = (a,b)$ is a solution, then $(C,S) = (b,a)$ is also a solution.
The selected pairs for each value of $\Omega$ are as follows:
\begin{itemize}
    \item $\Omega=2^{18}$: $(C,S) = (16, 173), (32,113), (48, 83), (83, 48), (113,32), (173,16)$. 
    \item $\Omega=2^{21}$: $(C,S) =(16,504), (32, 346), (64, 226),  (226,64), (346,32), (504,16)$.
    \item $\Omega=2^{27}$: $(C,S) = (128, 1386), (256, 904), (904, 256), (1386, 128)$.
\end{itemize}

\paragraph{(ii) SW-MLP.} 
For a fair comparison between the Mixers and SW-MLPs, we set the first layer of both models to the same per-patch FC structure.
We trained SW-MLPs with eight blocks, where the hidden layers of these MLPs share a common $\Omega = 2^{18}$ connectivity pattern. Following the per-patch fully connected layer, the feature matrix is vectorized and processed as a standard MLP with masked sparse connectivity.

For each freezing rate $1-p$, we determined the width $m$ of the hidden units using the equation:

\begin{align}
\Omega = m^2 p, \quad m = \sqrt{\Omega \over p }.
\end{align}

We set $1-p=0.1, 0.3, 0.5, 0.7, 0.9$, which correspond to $m=540, 612, 724, 935, 1619$, respectively.

\subsection{Figure~\ref{fig:SC}} \label{ssec:fig:sc:param}
We set the number of blocks $L$ and the other hyper-parameters as in  \cref{tab:hp-sc}.

\begin{table}[h]
    \centering
    \begin{tabular}{c||c|c|c|c|c|c|c|c}
       dataset  & $L$ & $\Omega$ & $\gamma$  & optimizer & init.\,lr & mini-batch & epoch & hard aug. \\
       \hline
       CIFAR10 & 12 & $2^{21}$ & None & AdamW & 0.02 & 1024 & 2000 & \checkmark
       \\

       CIFAR100 & 12 & $2^{21}$ & None & SGD & 0.1
       & 128 & 600 \\

       STL10  & 12 & $2^{18}$ & None & SGD & 0.1 & 128 & 2000 & \\

       ImageNet-1k & 8 & 290217984 & 4 & AdamW & 0.001 & 4096 & 300 & \checkmark
    \end{tabular}
    \caption{Hyper-parameters of models for normal/RP Mixers in \cref{fig:SC}.}
    \label{tab:hp-sc}
\end{table}

\paragraph{CIFAR-10, 100, STL-10.}
We utilized Tesla V100 GPUs and approximately 200 GPU hours for our experiments. We used a single GPU per each run.
We set $\Omega=2^{18}$ and used the same pairs $(S,C)$ satisfying \eqref{align:omega-non-gamma} as Sec. \ref{ssec:fig2}.

We set the initial learning rate to $0.1$. 
On CIFAR-10, we trained models 200 epochs,  600-epochs on CIFAR-100, and we did five trials with random seeds.  We trained models 2000-epochs on STL-10 with three random seeds.

\paragraph{ImageNet-1k.}
We utilized  Tesla V100 GPUs and approximately 4000 GPU hours for our experiments. for training MLP-Mixer and RP MLP-Mixer on ImageNet-1k; we used a GPU cluster of 32 nodes of 4 GPUs per node for each run.

We set the expansion factor $\gamma=4$ for both token-mixing MLP and channel-mixing MLP. 
We set $\Omega=290217984 =(768^2\cdot 196 + 768 \cdot 196^2)\gamma/2$  on a baseline $P=16, (S,C)=(196,768)$.
We sweep $P=7,8,14,16,28,32$ and set $C=3P^2$ and set $S$ so that it approximately satisfies the equation
\begin{align}
\Omega = {\gamma( CS^2 + SC^2 ) \over 2}.
\end{align}
For each setting, we did three trials with random seeds.

Training on the ImageNet-1k is based on the timm library~\cite{rw2019timm}.
We used AdamW with an initial learning rate of $10^{-3}$ and 300 epochs. We set the mini-batch size to 4096 and used data-parallel training with a batch size of 32 in each GPU.
We use the warm-up of with the warm-up learning rate $10^{-6}$ and the warm-up epoch $5$.
We used the cosine annealing of the learning rate with a minimum learning rate $10^{-5}$. We used the weight-decay 0.05.
We applied the random erasing in images with a ratio of 0.25. We also applied the random auto-augmentation with a policy rand-m9-mstd0.5-inc1.
We used the mix-up with $\alpha= 0.8$ and the cut-mix with $\alpha=1.0$ by switching them in probability 0.5. We used the label smoothing with $\varepsilon=0.1$.

\subsection{Figure \ref{fig:depth}}\label{ssec:fig5}
For our experiments in \cref{fig:depth}, we utilized Tesla V100 GPUs, with approximately 70 GPU hours utilized. We trained both S-Mixer and RP S-Mixer models on CIFAR-10, CIFAR-100, and STL-10 datasets.  We considered different numbers of blocks, specifically $L=4, 8, 12, 16, 20$. The values of $S$ and $C$ were fixed at 128. Each configuration was evaluated using three trials with different random seeds.

\section{Supplementary Experiments}

\subsection{Trainability of highly sparse weights}\label{sec:trainability}

\begin{figure}[h]
\centering
\includegraphics[width=0.3\textwidth]{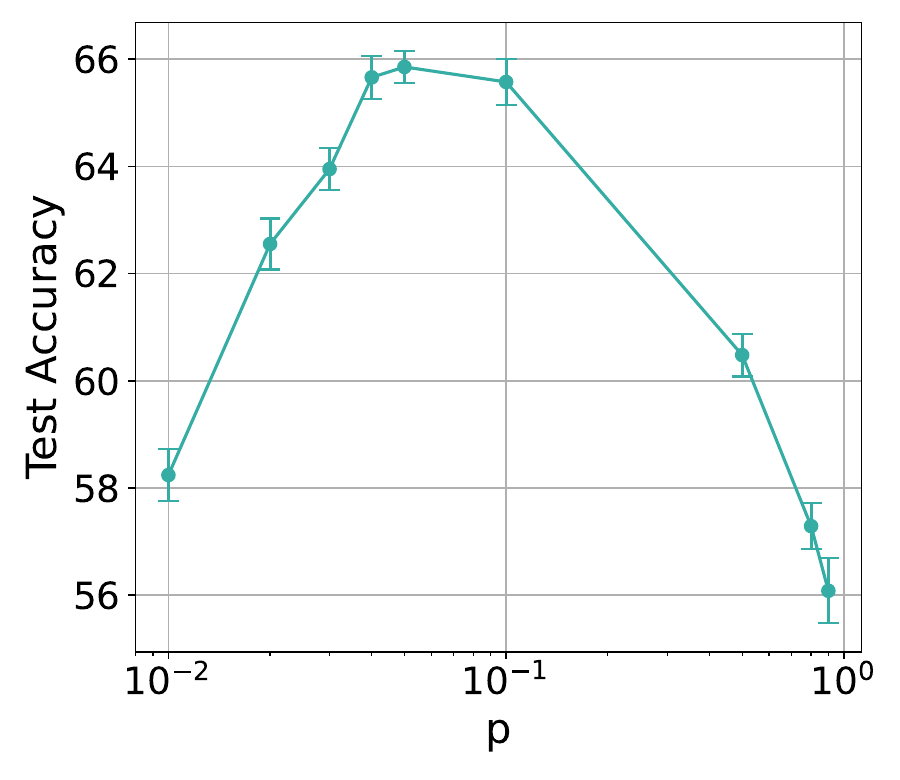}
\includegraphics[width=0.3\textwidth]{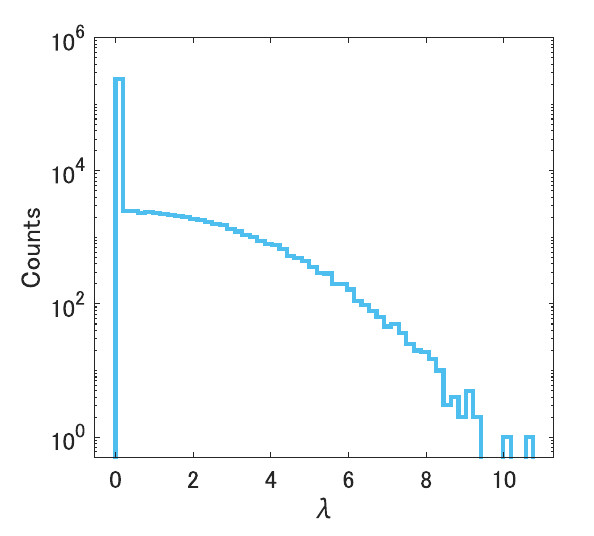}
\includegraphics[width=0.3\textwidth]{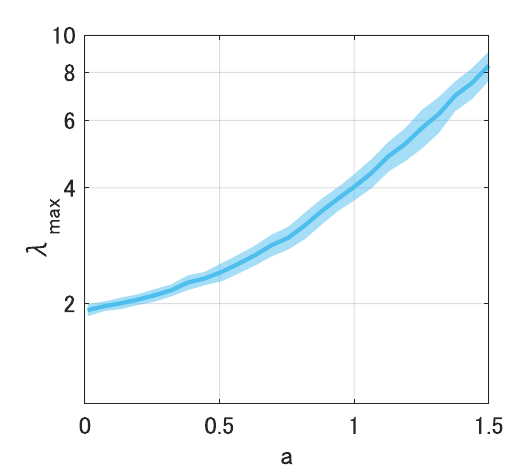}

\caption{On the trainability of SW-MLP. (Left) Trainability decreases as the sparsity $1-p$ becomes too high. We set $\gamma$ according to \cite{golubeva2021} under fixed $\Omega=2^{16}$. We performed five trials for each $p$ with random seeds. (Center) The singular value distribution of the sparse weight at random initialization. We set $a=1.5$, $\Omega=10^3$  and performed $50$ trials.
(Right) The largest eigenvalue monotonically increases as the sparsity increases. 
}
\label{fig:D2}
\end{figure}

\citet{golubeva2021} found that as the sparsity (width) increased to some extent, the generalization performance improved. They also reported that if the sparsity became too high, the generalization performance slightly decreased. 
They discussed that this decrease was caused by the deterioration of trainability, that is, it became difficult for the gradient descent to decrease the loss function. 
In fact, we confirmed their decrease of the performance in SW-MLP as is shown in \cref{fig:D2}(left).
In contrast, we hardly observed such a decrease in performance for the Mixers.
This seems rational because we can take an arbitrary small sparsity $1-p$ for the SW-MLP while it is lower-bounded for the Mixers as is described in Section 4. 

As a side note, we give here quantitative insight into the trainability from the perspective of the  singular value of the weight matrix. 
Some previous work reported that  the large singular values of weight matrices at random initialization cause the deterioration of trainability in deep neural networks \cite{bjorck2018understanding}. 
Following this line of study, let us consider  a random weight of SW-MLP. 
Set the width by $m = \Omega^{(1+a)/2}$ and the sparsity by $p = 1/\Omega^a$ with a constant $a>0$ and take a large $\Omega$ limit. We use this scaling because our interest is in the case where the expected number of weights, i.e., $ m^2 p=\Omega$, is independent of the scaling of $p$. 
We generate $ Z=   M  \odot  W $ where $W_{ij}\sim\mathcal{N}(0,1)$ and
$M$ is a static mask matrix whose entries are given by the Bernoulli distribution with probability $p$.
The singular value of the weight matrix $Z$ is equivalent to the square root of the eigenvalue of $Q=ZZ^\top$. Because we are uninterested in a trivial scale factor, we scaled the matrix $Q$ as $Q/c$ where $c$ denotes the average over the diagonal entries of $Q$. 
This makes the trace of $Q$, that is, the summation (or average) of eigenvalues, a constant independent of $a$. We computed the eigenvalues of $Q$ and obtained the singular values of $Z$ denoted by $\lambda$. 

As is shown in \cref{fig:D2}(center), the spectrum of the singular values peaked around zero but widely spread up to its edge. 
\cref{fig:D2}(right) demonstrates that the largest singular value becomes monotonically large for the increase of $a$. Because the larger singular value implies the lower trainability \cite{bjorck2018understanding}, this is consistent with the empirical observation of \cite{golubeva2021} and our \cref{fig:D2}(left). 

In contrast, the Mixers are unlikely to suffer from the large singular values as follows. 
Suppose S-Mixer with $S=C \gg 1$ for simplicity. Then, each layer of the effective MLP has $p=1/C$ which corresponds to the scaling index $a=1/3$ in SW-MLP. Actually, its singular value becomes further better than $a=1/3$, because the weight matrices of the normal and RP Mixers are structured: Consider the singular values of $Z= J_2 (I_C \otimes W) J_1$ with a $ C\times C$  random Gaussian matrix $W$ and permutation matrices $(J_1,J_2)$. Then, the singular values of $Z$ are equivalent to those of $W$, excluding duplication. Therefore, the singular values of the Mixers are determined only by the dense weight matrix $W$. Define $Q=WW^\top$.
Because the normalized matrix $Q/c$ obeys the Marchenko-Pastur law in the random matrix theory and its largest eigenvalue is given by 4 in the 
infinite $C$ limit \cite{bai2010spectral}. This means that the largest singular value of the normalized $W$ is 2 and corresponds to $a=0$ of SW-MLP (i.e., dense MLP) with the infinite $\Omega$ limit in \cref{fig:D2}(right). Thus, we can say that from the perspective of random weights, the trainability of the Mixers is expected to be better than that of SW-MLP.  

We can also extend our analysis to the models incorporating the expansion factor $\gamma$. For SW-MLP with MLP blocks, the expected number of weights is given by $\Omega = \gamma p m^2$. We just need to replace $p$ in the S-Mixer case to $\gamma p$ and the monotonic increase of the largest singular value appears as well.  
For the MLP-Mixer, its normalized $W$ is a $\gamma C \times C$ matrix. According to the Marchenko-Pastur law for rectangular random matrices, as $C \to \infty$, the largest singular value approaches a constant value of $1 + \sqrt{\gamma}$. This corresponds to the singular value of $a=0$ in the corresponding SW-MLP, and the result is similar as in the S-Mixer.

Note that the effective width of the mixing layers is sufficiently large but still has an upper bound (\ref{max_width}). It satisfies 
\begin{equation}
(\sqrt{1+8\Omega/\gamma}-1)/2  \leq  m \leq (\Omega/\gamma)^{2/3},
\end{equation}
where the equality of the lower bound holds for $S=1$ or 
$C=1$. In contrast, for SW-MLP, we have no upper bound and only the lower bound 
$\sqrt{\Omega} \leq m$, where this equality holds for a dense layer.  We can consider an arbitrarily small $p$ and a large $m$ for a fixed $\Omega$ if we neglect the issue of memory \citep{golubeva2021}.
\citet{golubeva2021} reported that extremely small $p$ can cause a decrease in test accuracy owing to the deterioration of trainability. We observed a similar deterioration for the SW-MLP, as shown in this section, but not for the Mixers. This is expected because $m$ is upper-bounded in the Mixers and the trainability is less damaged than that of the SW-MLP with high sparsity.

\subsection{MLP-Mixer with increasing width}\label{sec:increasing_width_mlpmixer_appendix}

\begin{figure}[h]
\centering
\includegraphics[width=\textwidth]{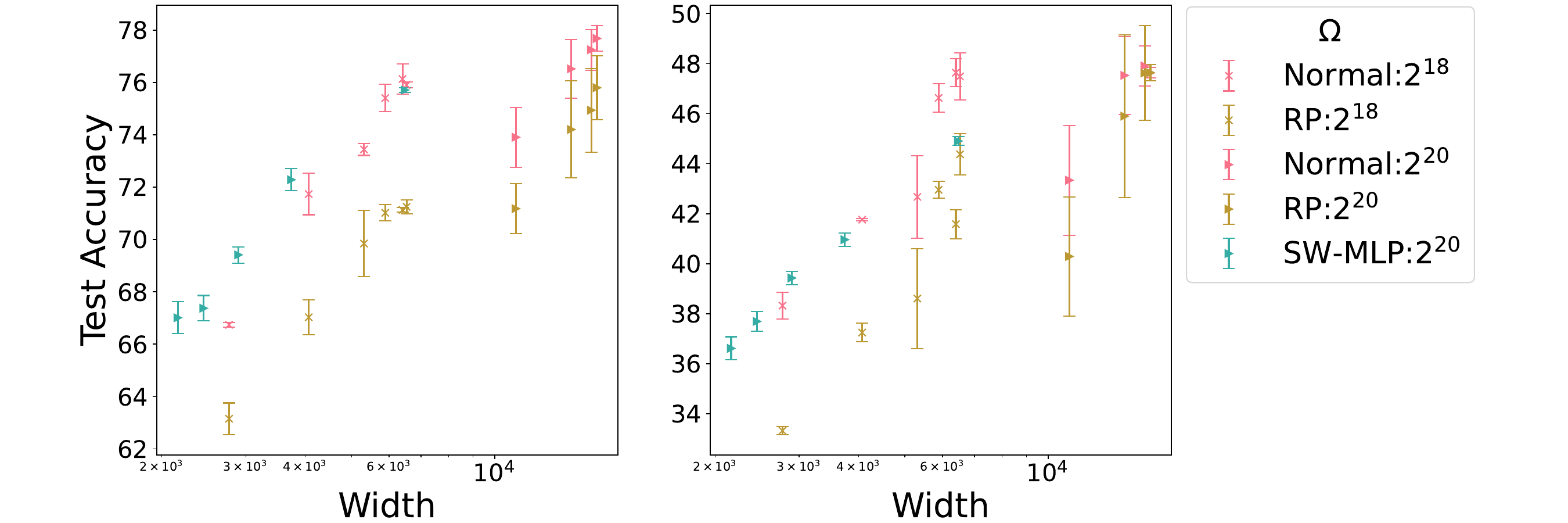}

\caption{ Test accuracy improves as the effective width increases.  MLP-Mixer, RP-MLP-Mixer, and SW-MLP with $\gamma=4$ on (Left) CIFAR-10, (Right) CIFAR-100.}

\label{fig:D1}
\end{figure}

\cref{fig:D1} shows the test accuracy improves as the width increases on the models SW-MLP and normal and RP MLP-Mixer even if the expansion factor $\gamma$ is incorporated in the models. We set the expansion factor $\gamma=4$.   We set the initial learning rate to be 0.1.
For normal and RP MLP-Mixer, we set $C=16,32,48,64, 83,113,173$ and determined $S$ by combinations of $C$ and $\Omega= 2^{18}, 2^{20}$.
For SW-MLP, we set $p=0.1,0.3,0.5,0.7,0.9$ and set the width by $m = \sqrt{\Omega/\gamma p}$. 

\subsection{Increasing expanding factor}
\label{sec:experiments-3}

\begin{figure*}[t]
\centering
\includegraphics[width=0.48\textwidth]{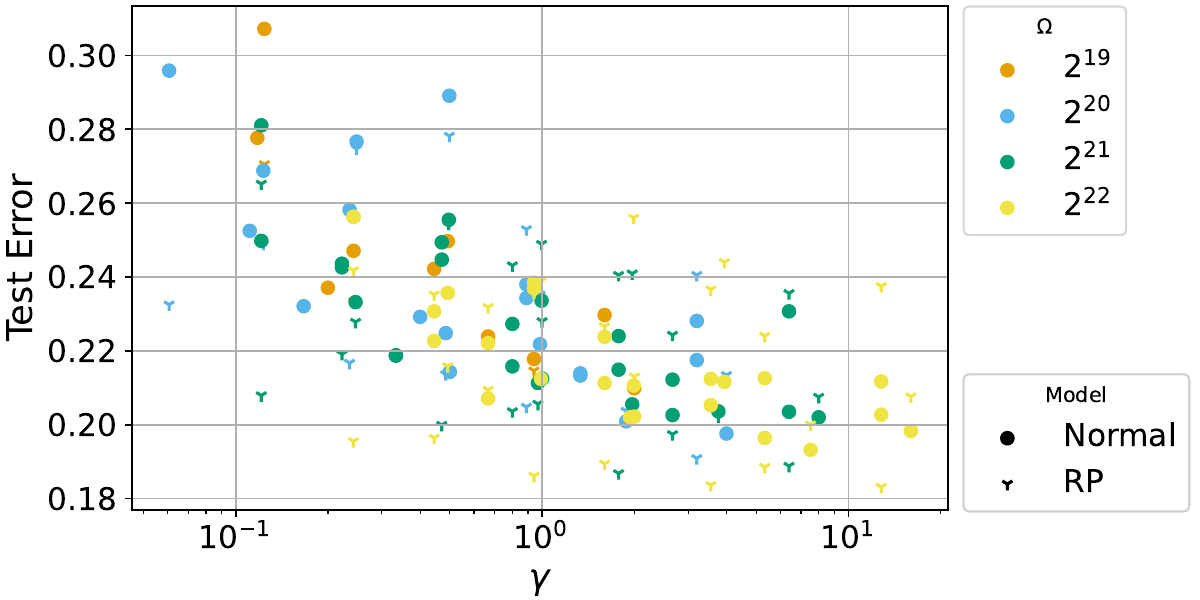}
\includegraphics[width=0.33\textwidth]{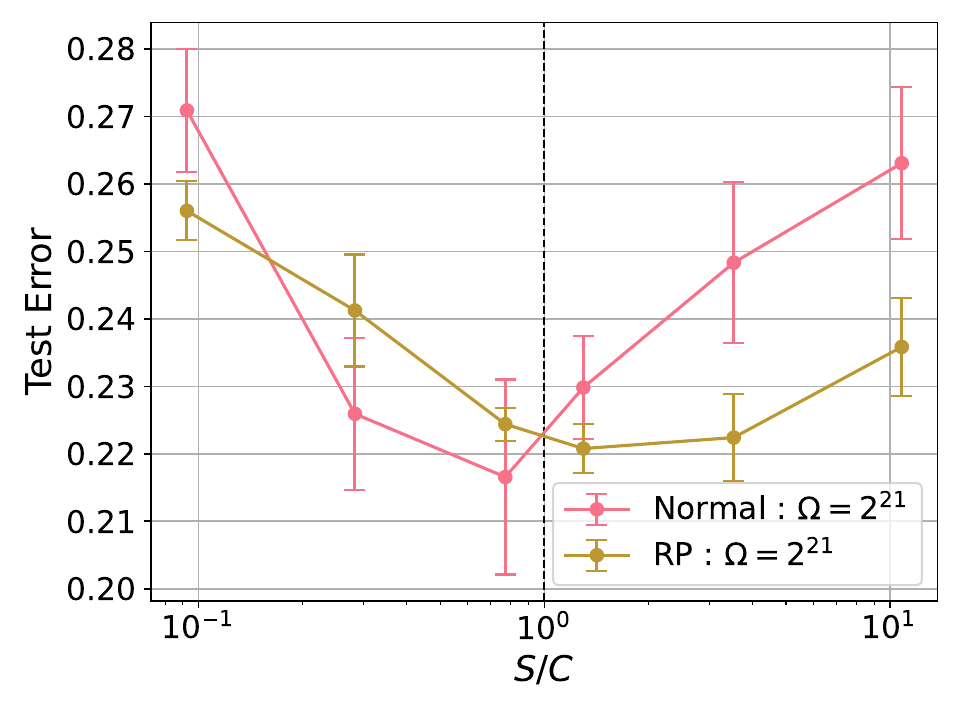}

\caption{(Left) Increasing expansion factor $\gamma$ improved the test error in normal and RP MLP-Mixers.  (Right) The lowest error is achieved around $C=S$ with fixed $m=4096$.}
\label{fig:expand}
\end{figure*}

The effective MLP expression of the MLP-Mixer (\ref{eq:effMLPmixer}) has two widths: $m=SC$ and $\gamma SC$. As both are proportional to $SC$, we have focused on changing $SC$ and fixed $\gamma$ so far.  
Here, we consider a complementary setting, that is, changing $\gamma$ with fixed $m=SC$. By substituting $S=m/C$ into  (\ref{eq:omega}), we obtain $\gamma = 2\Omega/(m(C+m/C)).$
Similar to $m$ of $C$ shown in Fig.\,\ref{fig:e}, this $\gamma$ is a single-peak function of $C$ and takes its maximum as 
\begin{equation}
C^* =S^* = \sqrt{m}, \quad \max_{S,C} \gamma= \Omega/(m\sqrt{m}). \label{eq:max_gamma} 
\end{equation}
Fig.\,\ref{fig:expand}(left) confirms that increasing the width ($\gamma$) leads to performance improvement as is expected from Hypothesis\ref{hyp}. We trained normal and RP MLP-Mixers with various $\gamma$ in a realistic range. We plotted some cases of fixed $\Omega$ under the same $m$. 
Fig.\,\ref{fig:expand}(right) shows the test accuracy maximized around $C=S$
as is expected from  (\ref{eq:max_gamma}).

\subsection{Dependence on depth}\label{sec:depth}

As shown in Figs.\ref{fig:width}-\ref{fig:expand}, both the normal and RP Mixers exhibited similar tendencies for a fixed depth. Fig.\,\ref{fig:depth} confirms that by increasing the depth, i.e., the number of blocks, RP S-Mixers can even become comparable to
the normal ones or better than them in some cases. First, we observed that, when the depth was limited, the RP Mixers were inferior to the normal Mixers in most cases. As we increased the depth, we observed that in some cases, overfitting occurred for the normal Mixer, but not for the RP one (see also the training and test losses shown in Section D.6). In such cases, the results of the RP Mixers were comparable (in Figs.\,\ref{fig:depth}(left, right)) or better (in Fig.\,\ref{fig:depth}(center)). Although RP Mixers are not necessarily better than normal ones, it is intriguing that even RP Mixers defined by a random structure can compete with normal Mixers.

The random permutation in RP-Mixer initially selects and fixes {\it the size of receptive field} (i.e., the number of 
input pixels flowing into a neuron in each upper layer through weights) 
at a rate of $1/\sqrt{m}$. Therefore, the training becomes difficult in shallow networks. However, in the RP-Mixer, since an independent random permutation is chosen for each layer, the receptive field expands as the number of layers increases. From this, the performance of the RP-Mixer improves as layers are added. The normal MLP-Mixer performs well even with fewer layers because the receptive field and diagonal-block structure are aligned.

\subsection{Replacement of $J_c$ in specific layers}\label{sec:replacement}
\begin{figure}[t]
\centering
\includegraphics[width=\textwidth]{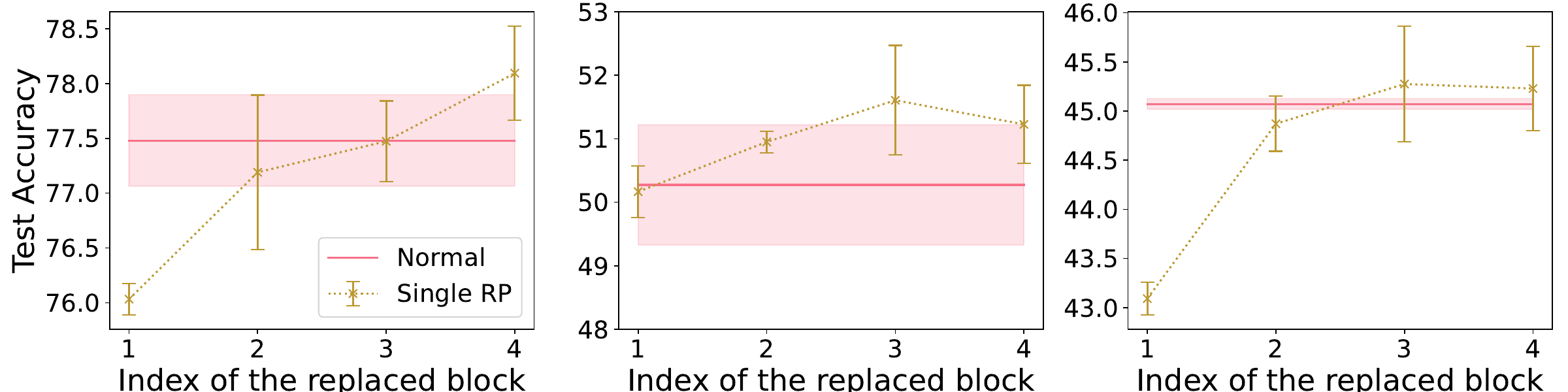}

\caption{Replacing a single block of the normal Mixer with a corresponding RP block clarifies that the upstream layers are functionally commutative with the RP block.  (Left) CIFAR-10, (Center) CIFAR-100, (Right) STL-10. We set $C=S=128$. }
\label{fig:replace}
\end{figure}

\cref{fig:replace} provides more detailed insight into the case where the depth is limited and RP Mixers perform worse than normal Mixers. We investigated special S-Mixers whose $l$-th block was replaced with its RP counterpart while the other layers remained the same. Interestingly, when the accuracy deterioration apparently appears (e.g., cases of CIFAR-10 and STL-10 in \cref{fig:depth}), this deterioration is attributed to the first block. This seems rational because the neighboring image patches are likely to be correlated, which makes the input units to the first token mixing correlated. Although the usual mixing weights can reflect such neighboring structures, RP Mixers randomly choose tokens and may lose the neighboring structure specific to the images. However, as the depth increases, the token mixing layers can merge all the tokens, which is consistent with the increase in the accuracy of the RP Mixers, as confirmed in \cref{fig:depth}.
Thus, we conclude that the RP and normal mixing layers have almost the same inductive bias, especially, in the upstream layers.

We utilized Tesla V100 GPUs and approximately 10 GPU hours for our experiments.
Consider the S-Mixer architecture consisting of four blocks and $S=C=64$. In this study, we trained a modified version of the S-Mixer architecture by replacing one of the four blocks with a block that incorporates random permutation. The training was conducted on CIFAR-10, CIFAR-100, and STL-10 datasets. The optimizer and training settings used in this experiment were consistent with those described in Section \ref{ssec:fig5}.

\subsection{Remark on input patch size}

\begin{figure}[t]
\centering
\includegraphics[width=0.5\textwidth] {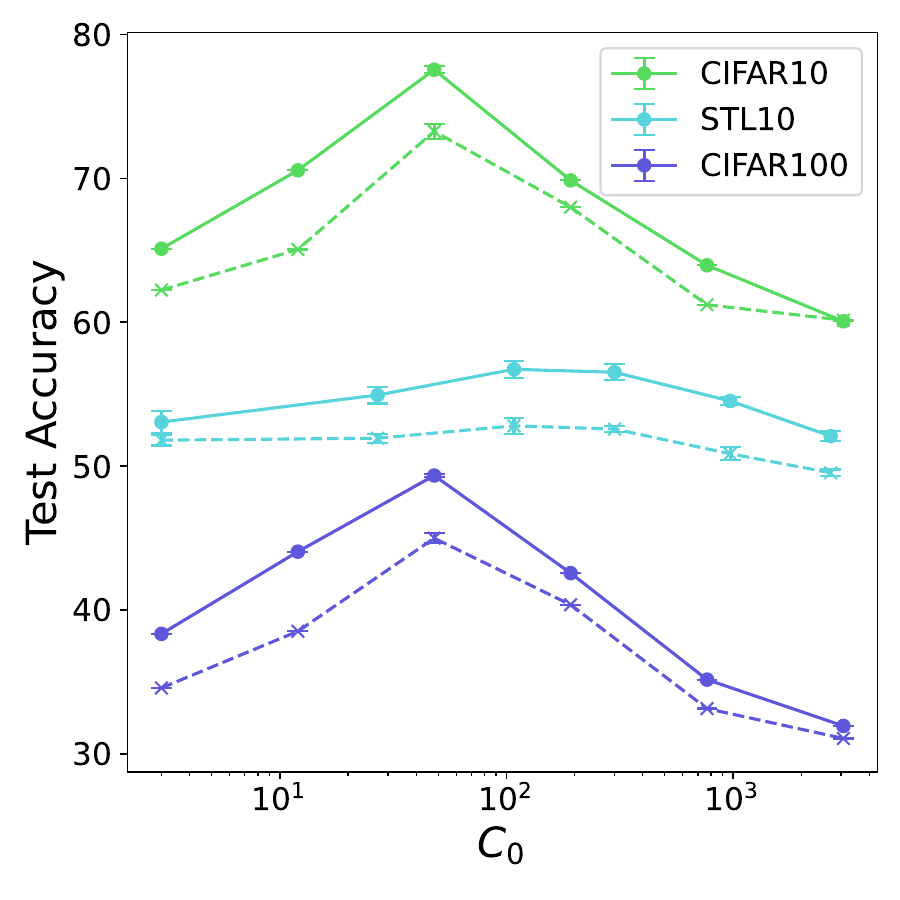}
\vspace{-18pt}
\caption{Dependence on $C_0$}
\label{fig:patchsize}
\vspace{-0.15in}
\end{figure}

In this study, we focused on changing the size of the mixing layers and fixed the input token and channel size ($S_0, C_0$). In other words, the patch size $P$, satisfying $C_0 =3 P^2$, is fixed.
While we observe that our experimental results hold regardless of the patch size, 
one naive question is whether there is any optimal patch size for achieving the highest accuracy.
Although this is beyond the scope of this study, we show 
the performance depending on $C_0$ in 
 \cref{fig:patchsize} as a side note. 
 The number of mixing layers is fixed at $C=S=64$. We observed that the optimal $C_0$ depended on data; $C_0=48$ $(S_0=64)$ for CIFAR-10 and 100, and $C_0=108$ $(S_0=225)$ for STL-10. Note that the dimension of an input image is $S_0 C_0 = 3,072$ for CIFAR datasets and $24,300$ for STL-10.
 It would be rational that the optimal patch size depends on the detailed information of data.

We utilized Tesla V100 GPUs and approximately 30 GPU hours for our experiments.
We conducted training experiments on CIFAR-10, CIFAR-100, and STL-10 datasets using both S-Mixer and RP S-Mixer architectures with $S=C=64$,  along with four blocks. For optimization, we set the initial learning rate to be 0.1. 

For CIFAR-10 and CIFAR-100, we trained the models for 200 epochs and evaluated them with different patch sizes ($P=1,2,4,8,16,32$). We performed three trials with different random seeds for each setting. 
On the STL-10 dataset, we resized the images to $90 \times 90 \times 3$ and trained the models for 400 epochs. We varied the patch size ($P=1,3,6,10,18,30$) and performed five trials with different random seeds for each setting.

\subsection{RP can prevent the overfitting}

\begin{figure}[h]
\centering
\includegraphics[width=0.49\textwidth]{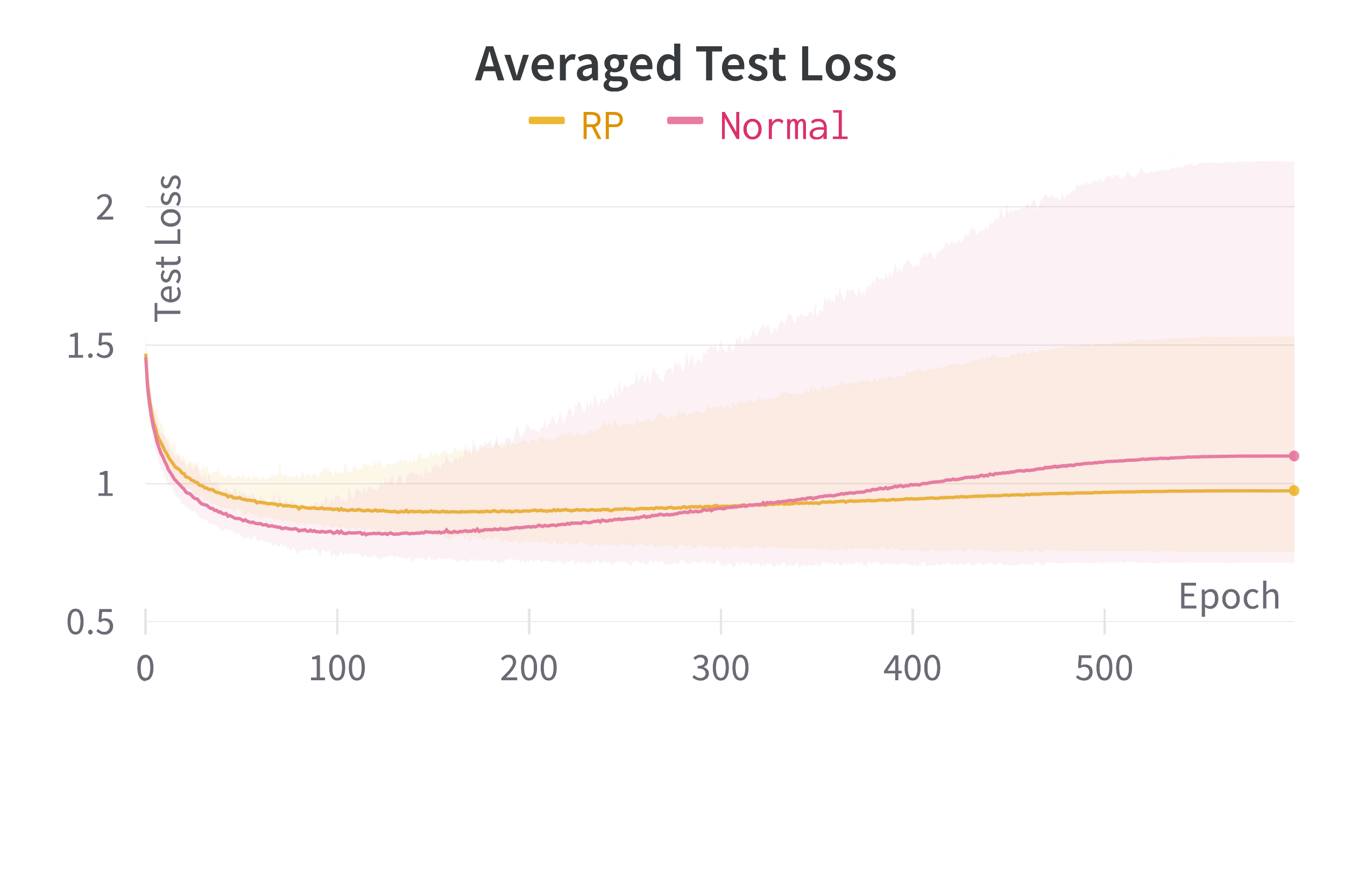}
\includegraphics[width=0.49\textwidth]{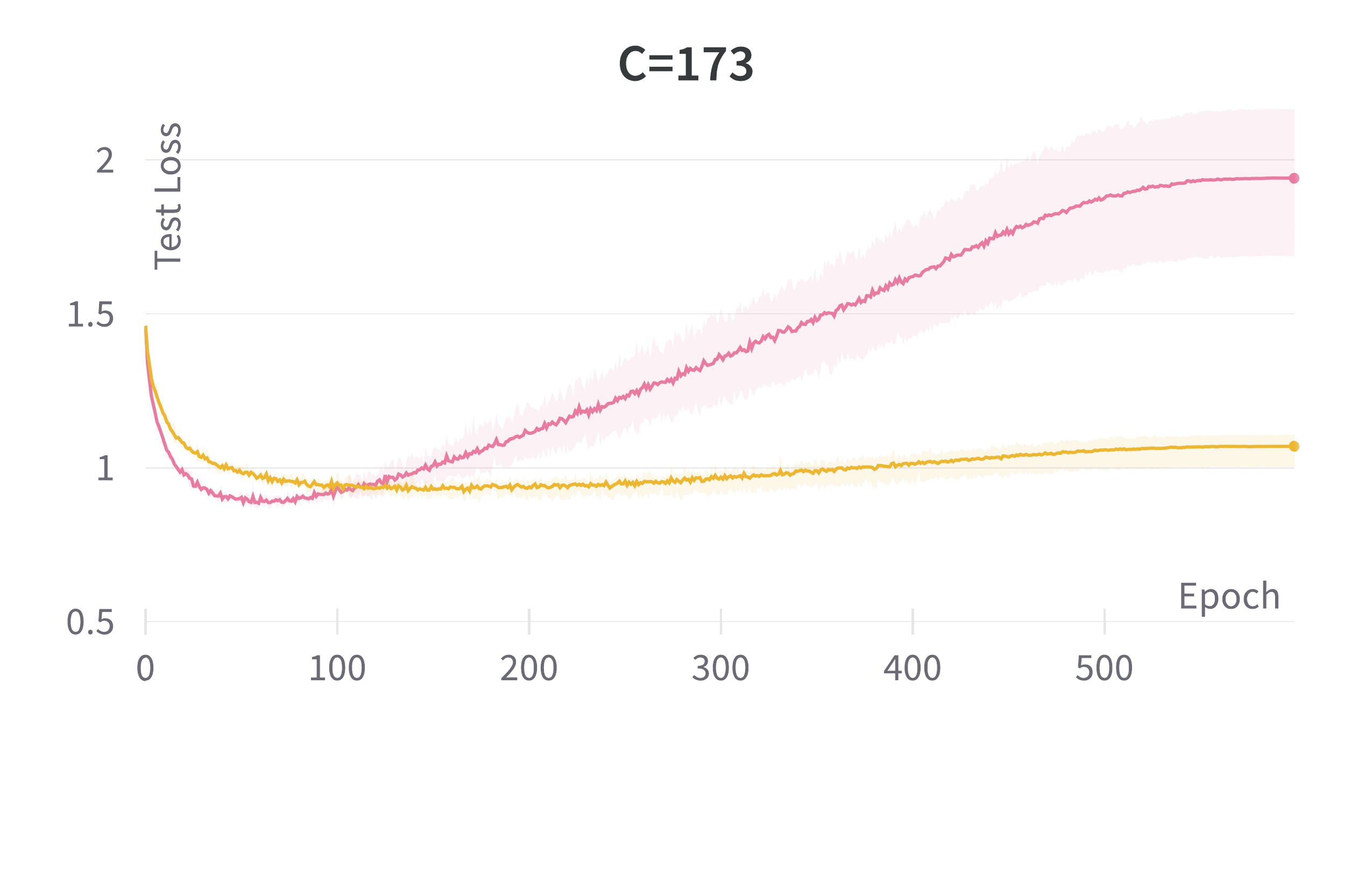}

\caption{
(Left) The average test loss curves are shown for $C=16,32,64,114,173$ and five trials with different random seeds. The models used in this experiment were Normal and RP S-Mixers trained on CIFAR-10 with $L=8$ for 600 epochs. The initial learning rate was set to 0.1.
(Right) The test loss curve for $C=173$ represents the worst case of overfitting. The shaded area in both figures represents the range between the maximum and minimum values.}
\label{fig:D3}

\end{figure}

In \cref{fig:D3}, we explored several values of $C$ fixed $\Omega$, the normal model shows overfitting of more than twice the magnitude compared to the RP model, especially $C$ is the largest one in the exploring range. In this case, $S$  takes the smallest value among the explored values.  This suggests that RP has a regularization effect beyond the token-mixing side and affects the channel-mixing side, particularly when $C$ is large. 

\begin{figure}[ht]
\centering
\includegraphics[width=0.49\textwidth]{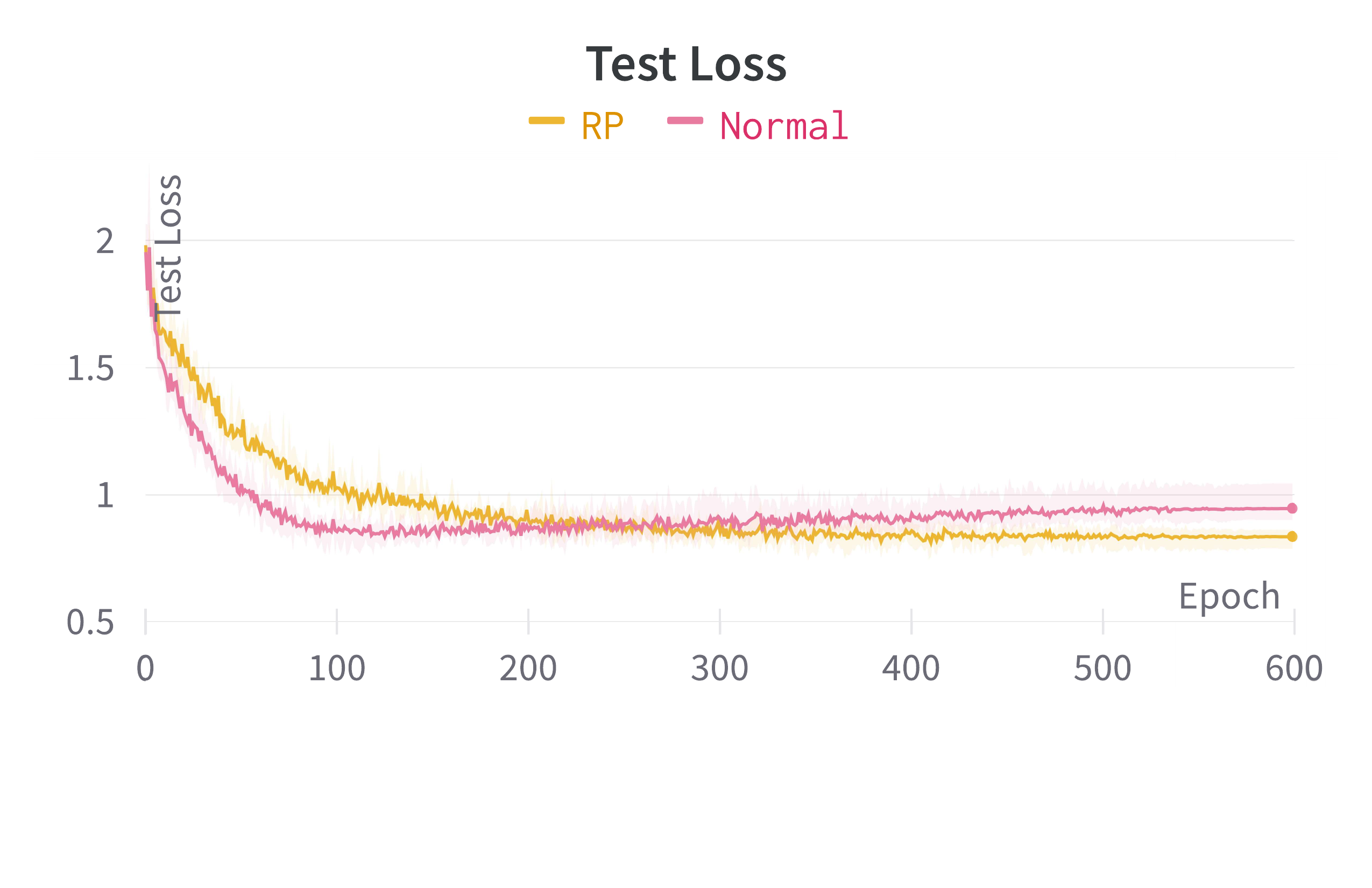}
\includegraphics[width=0.49\textwidth]{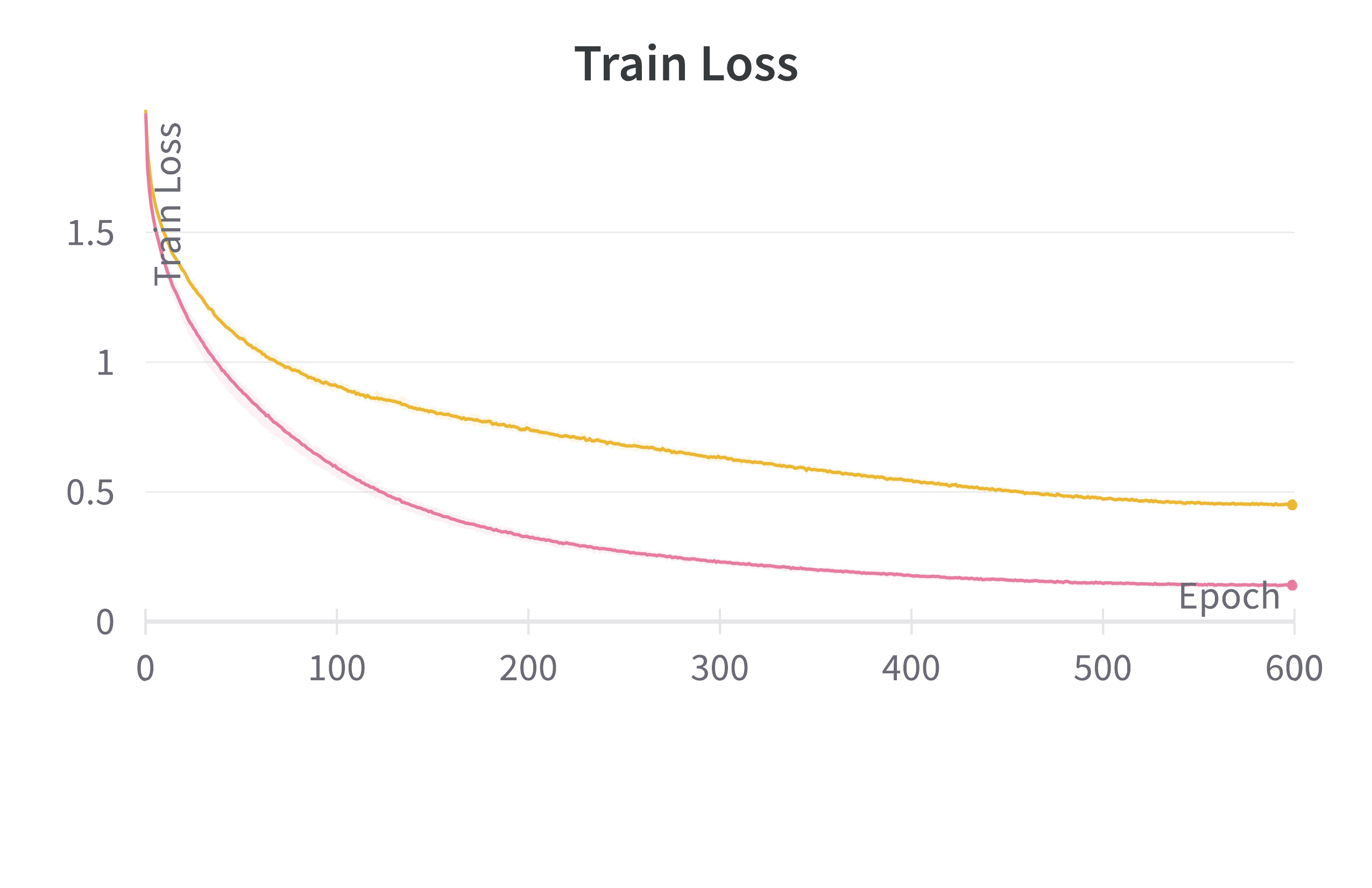}

\caption{  (Left) Average of test loss curves.  (Right) Average train loss with $C=173$.  In both figures, the area shows max and min values.  We trained normal and RP MLP-Mixer with $\gamma=4$ with an initial learning rate of 0.1 and a mini-batch size of 256  for 600 epochs.  The results are average of five trials.   The shaded area in the figure represents the range of values between the maximum and minimum values.}
\label{fig:D3-autoaug}
\end{figure}

To mitigate overfitting, additional augmentation (auto-augmentation, based on \cite{cubuk2019autoaugment}) was applied to the dataset, and the model was switched from S-Mixer to MLP-Mixer ($\gamma=4$) due to the observed slower decrease in training loss for S-Mixer in \cref{fig:D3-autoaug}. RP S-Mixer outperformed S-Mixer in terms of test loss for $C=173$, indicating that RP still provides overfitting prevention even with relatively strong data augmentations.

\begin{figure}[ht]
\centering
\includegraphics[width=0.45\textwidth]{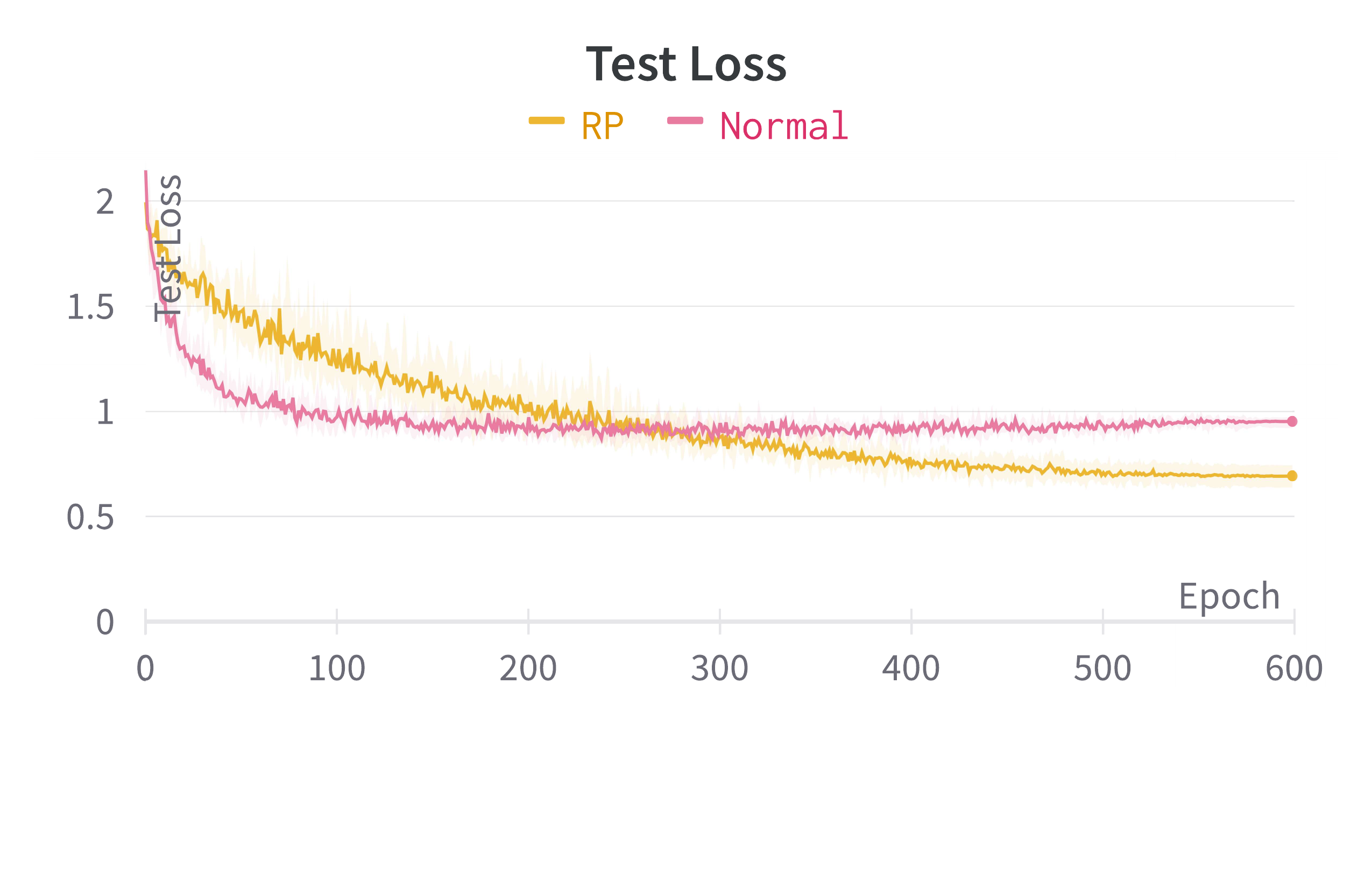}
\includegraphics[width=0.45\textwidth]{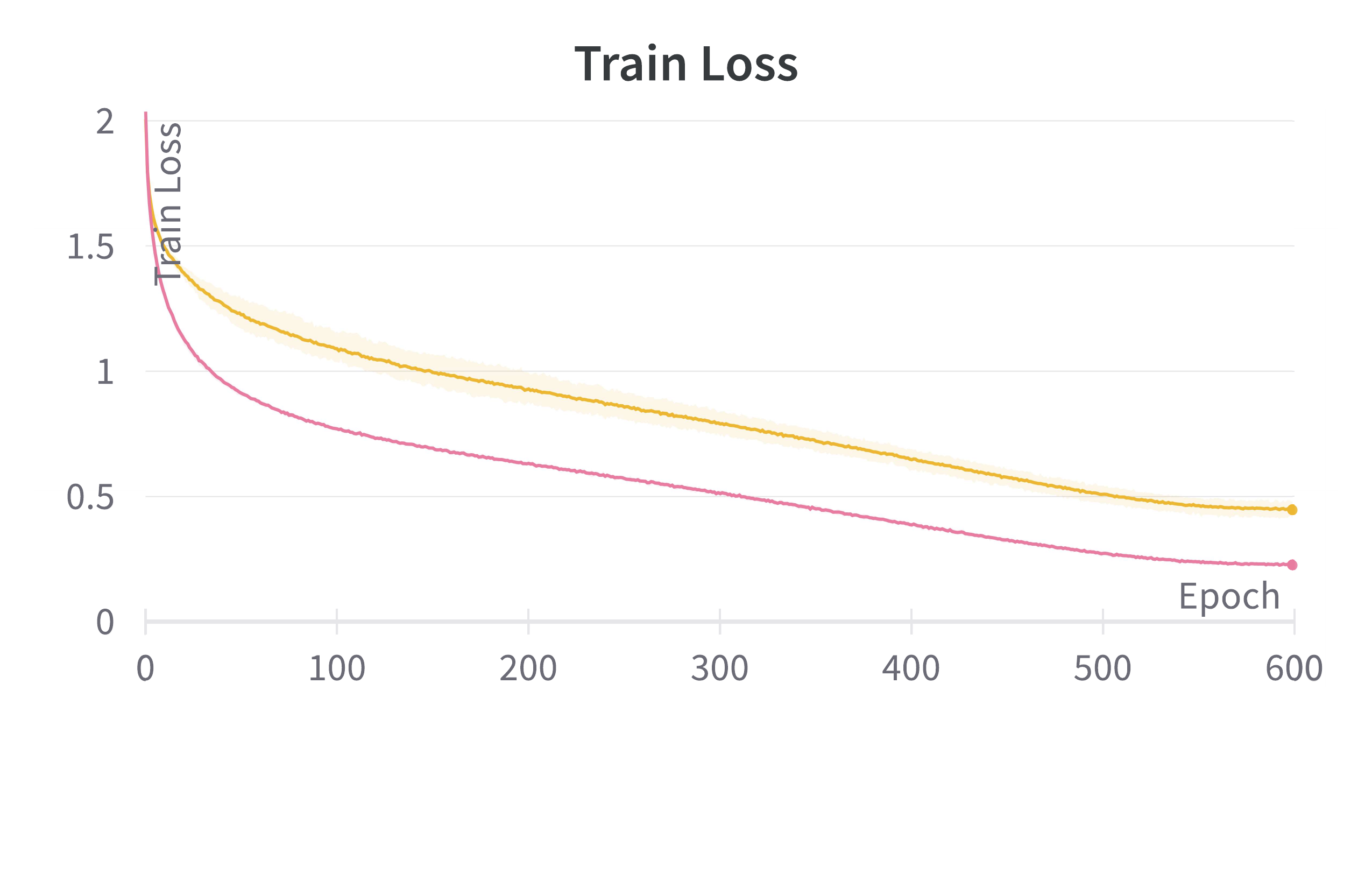}
\includegraphics[width=0.45\textwidth]{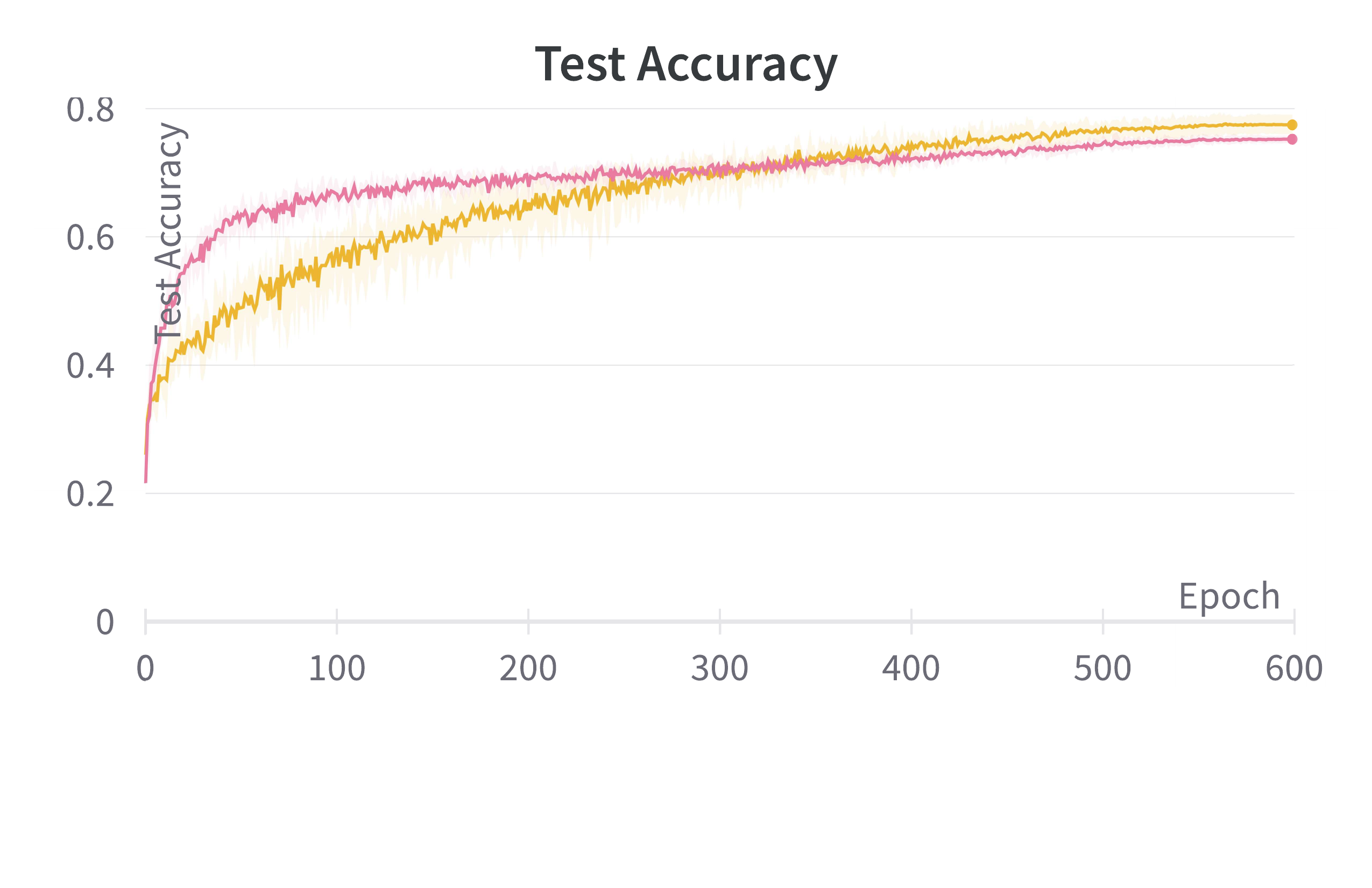}

\caption{ Results of training with AdamW and auto-augmentation with $S=C=64$. The RP MLP-Mixer exceeded the results of the normal one in test loss and test accuracy. (Left) Average of test loss curves. (Right) Average train loss. (Lower) Test Accuracy curves.   We set the initial learning rate to be 0.01 with a mini-batch size of 256 and 600 epochs.   In all figures, the results are average of five trials and the area shows the range between the maximum and minimum values.}
\label{fig:D3-adamw}
\end{figure}

\cref{fig:D3-autoaug} (right) illustrates that RP did not reach the relatively small training loss as the normal model.
To address this, SGD was replaced with AdamW as the optimizer, with a reduced initial learning rate (lr $=0.01$) due to the instability observed with lr $=0.1$ in \cref{fig:D3-adamw}. This resulted in reduced overfitting in the $C > S$ region, and RP performed exceptionally well compared to the normal model for $C=S=64$.
In \cref{fig:D3-adamw}, neither the normal nor RP models exhibited a significant increase in test loss for $C=S$. However, while the normal model's test loss plateaued, the RP model continued to decrease its test loss, eventually surpassing the normal model in terms of test accuracy. This highlights the potential of RP to outperform the normal model with a choice of optimization such as AdamW.

\subsection{Visualization of how random permutation breaks the structure}

\begin{figure}[th]
    \centering
    \includegraphics[width=0.7\linewidth]{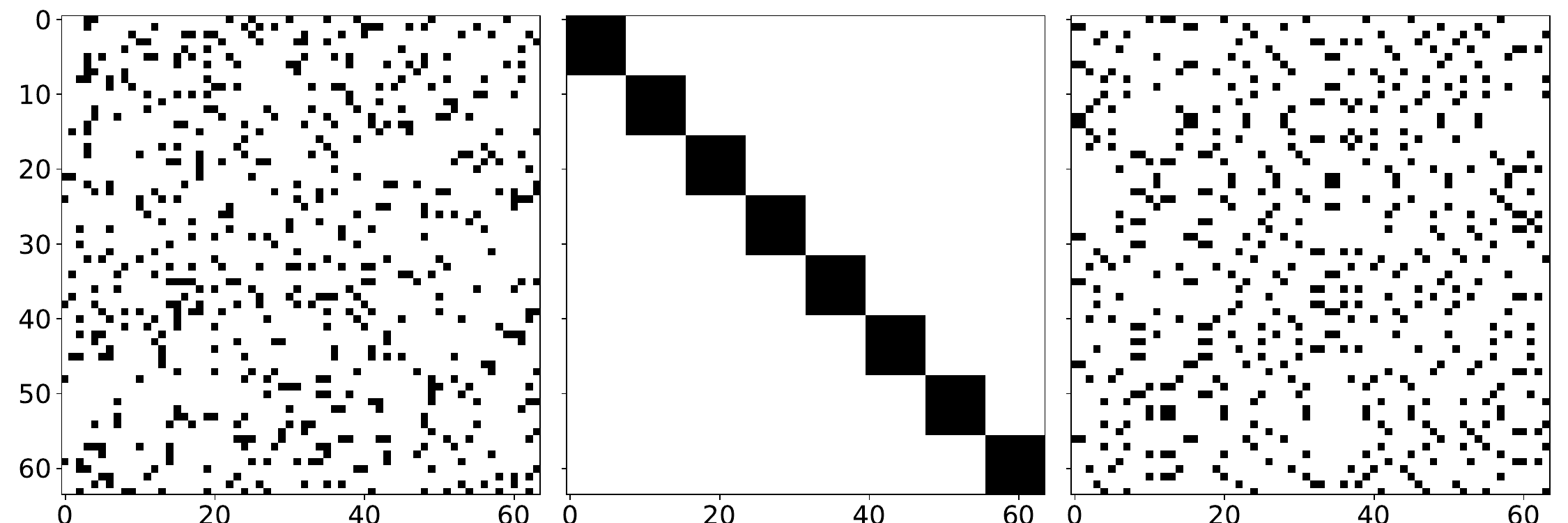}
     \caption{Samples of weight matrices at initialization. (Left) $M \odot A$, (center)$I_C \otimes W$, (right)$J_1(I_C \otimes W)J_2$, $C=S=8, m=CS, p=1/S$, and $J_1$ and $J_2$ are independent random permutation matrices.
    Each sample is a realization of one trial.}
    \label{fig:mat_visual}
\end{figure}

Observing the differences between RP-Mixer, MLP-Mixer, and SW-MLP from the perspective of weights. \cref{fig:mat_visual} shows three types of models where components can only be 0 or 1. In RP-Mixer, it is evident that the block structure of MLP-Mixer has been disrupted.

\subsection{Computational Costs}\label{sec:cost}
Table~\ref{tab:compuation} shows the computational resource of SW-MLPs and RP-Mixers. In the setting for ImageNet, SW-MLP requires huge memory and runtime, whereas RP-Mixer needs $10^3$ to $10^6$ times less. Note that the spacial complexity of the SW-MLP (resp.\,the RP-Mixer) is $O(m^2)=O(S^2C^2)$ (resp.\,$O(C^2 + S^2)$) as $S,C \to \infty$. Therefore, the RP-Mixer is more memory-efficient and computationally undemanding than the SW-MLP.
\begin{table*}[ht]
    \centering
    \begin{tabular}{c|c|c|c}
        Model & Memory(CIFAR/ImageNet) & FLOPs(CIFAR/ImageNet) & Runtime(CIFAR)\\
        \hline
        SW-MLP & 3.22GB / 23.2TB & 805M / 5.80T& $28.2(\pm 0.00)$ s/epoch\\
        RP-Mixer &  3.94MB / 26.3MB  & 12.5M / 4.06G& $6.41(\pm 0.01)$ s/epoch\\
        MLP-Mixer &  3.93MB / 26.3MB  & 12.5M / 4.06G& $6.08(\pm 0.26)$ s/epoch
    \end{tabular}
    \caption{Comparison on memory requirements, FLOPs(floating point operations), and averaged runtime. For the three models, we set $S=256, C=588, L=8, \gamma=4$ for ImageNet and $S=64, C=48,L=3, \gamma=2$ for CIFAR. }
    \label{tab:compuation}
\end{table*}

\end{document}